\documentclass{article} %
\usepackage{iclr/iclr2021_conference,times}
\usepackage{times}
\usepackage{latexsym}

\usepackage[T1]{fontenc}

\usepackage[utf8]{inputenc}
\usepackage{url}

\usepackage[utf8]{inputenc}
\usepackage{graphicx}
\usepackage{xcolor}

\usepackage{booktabs}
\usepackage{tabulary}

\usepackage{graphicx}

\usepackage{caption}
\usepackage{subcaption}

\usepackage{amssymb}
\usepackage{amsmath}

\usepackage{textcomp}

\usepackage{contour}
\usepackage[normalem]{ulem}

\contourlength{0.8pt}

\usepackage[normalem]{ulem}  %
\newif\ifcomments
\commentstrue

\definecolor{darkblue}{rgb}{0, 0, 0.5}

\usepackage[colorlinks=true,citecolor=darkblue, linkcolor=darkblue, urlcolor=darkblue]{hyperref}
\makeatletter
\renewcommand{\sectionautorefname}{\S\@gobble}
\renewcommand{\sectionautorefname}{\S\@gobble}
\renewcommand{\subsectionautorefname}{\S\@gobble}
\renewcommand{\sectionautorefname}{\S\@gobble}
\renewcommand{\subsectionautorefname}{\S\@gobble}
\makeatother

\usepackage{cleveref}

\usepackage{csquotes}

\usepackage{appendix}
\usepackage{etoolbox}
\makeatletter
\patchcmd{\hyper@makecurrent}{%
    \ifx\Hy@param\Hy@chapterstring
        \let\Hy@param\Hy@chapapp
    \fi
}{%
    \iftoggle{inappendix}{%
        \@checkappendixparam{chapter}%
        \@checkappendixparam{section}%
        \@checkappendixparam{subsection}%
        \@checkappendixparam{subsubsection}%
        \@checkappendixparam{paragraph}%
        \@checkappendixparam{subparagraph}%
    }{}%
}{}{\errmessage{failed to patch}}

\newcommand*{\@checkappendixparam}[1]{%
    \def\@checkappendixparamtmp{#1}%
    \ifx\Hy@param\@checkappendixparamtmp
        \let\Hy@param\Hy@appendixstring
    \fi
}
\makeatletter

\newtoggle{inappendix}
\togglefalse{inappendix}

\apptocmd{\appendix}{\toggletrue{inappendix}}{}{\errmessage{failed to patch}}
\apptocmd{\subappendices}{\toggletrue{inappendix}}{}{\errmessage{failed to patch}}

\definecolor{affirmgreen}{rgb}{0,0.5,0}
\definecolor{negatered}{rgb}{1,0,0}
\newcommand*{\affirm}[1]{\textcolor{affirmgreen}{\textbf{#1}}}
\newcommand*{\negate}[1]{\textcolor{negatered}{\textbf{#1}}}

\usepackage{amsmath,amsfonts,bm}

\def\eqref#1{equation~\ref{#1}}

\def\1{\bm{1}}

\DeclareMathAlphabet{\mathsfit}{\encodingdefault}{\sfdefault}{m}{sl}
\SetMathAlphabet{\mathsfit}{bold}{\encodingdefault}{\sfdefault}{bx}{n}

\usepackage{url}
\usepackage{xcolor}
\usepackage{hyperref}

\definecolor{agree}{HTML}{1770ab}
\definecolor{weaklyagree}{HTML}{94c6da}
\definecolor{weaklydisagree}{HTML}{f3a583}
\definecolor{disagree}{HTML}{c30d24}

\newcommand*{\Q}{\ensuremath{\mathcal{Q}}}

\newcommand*{\answerstyle}[1]{\textsc{#1}}
\newcommand*{\Agree}{\textcolor{agree}{\answerstyle{Agree}}}
\newcommand*{\WeaklyAgree}{\textcolor{weaklyagree}{\answerstyle{Weakly agree}}}
\newcommand*{\WeaklyDisagree}{\textcolor{weaklydisagree}{\answerstyle{Weakly disagree}}}
\newcommand*{\Disagree}{\textcolor{disagree}{\answerstyle{Disagree}}}

\ifcomments
    \newcommand*{\TODO}[1]{\textcolor{red}{[TODO: #1]}}
    \newcommand*{\tocite}[1]{(\textcolor{blue}{#1})}
    \newcommand*{\tocitep}[1]{(\textcolor{blue}{#1})}
    \newcommand*{\tocitet}[1]{\textcolor{blue}{#1}}

    \newcommand*{\maybedelete}[1]{\textcolor{red}{\sout{#1}}}
    
    \newcommand*{\julian}[1]{\textcolor{orange}{[JM: #1]}}
    
\else
    \newcommand*{\julian}[1]{}
    \newcommand*{\TODO}[1]{}
    \newcommand*{\tocite}[1]{}
    \newcommand*{\tocitep}[1]{}
    \newcommand*{\tocitet}[1]{}

    \newcommand*{\maybedelete}[1]{}
    
\fi

\title{%
What Do NLP Researchers Believe? \\
Results of the NLP Community Metasurvey}

\renewcommand\sup[1]{\rlap{$^{#1}$}}

\author{Julian Michael\sup{1,2},~~~~Ari Holtzman\sup{1},~~Alicia Parrish\sup{4},~~Aaron Mueller\sup{5},~~Alex Wang\sup{3},\\ \textbf{Angelica Chen\sup{2},~~Divyam Madaan\sup{3},~~Nikita Nangia\sup{2},} \\ \textbf{Richard Yuanzhe Pang\sup{3},~~Jason Phang\sup{2},}~~and \\ \textbf{Samuel R. Bowman\sup{2,3,4}}
\\
\textsuperscript{1}Paul G. Allen School for Computer Science \& Engineering, University of Washington \\
\textsuperscript{2}Center for Data Science, New York University \\
\textsuperscript{3}Courant Institute of Mathematical Sciences, New York University \\
\textsuperscript{4}Department of Linguistics, New York University \\
\textsuperscript{5}Department of Computer Science, Johns Hopkins University \\
\texttt{nlp-metasurvey-admin@nyu.edu}
}

\iclrfinalcopy %
\begin{document}

\maketitle

\begin{abstract}
We present the results of the NLP Community Metasurvey.
Run from May to June 2022,
the survey elicited opinions on controversial issues, including industry influence in the field, concerns about AGI, and ethics.
Our results put concrete numbers to several controversies: For example, respondents are split almost exactly in half on questions about the importance of artificial general intelligence, whether language models understand language, and the necessity of linguistic structure and inductive bias for solving NLP problems.
In addition, the survey posed \textit{meta}-questions, asking respondents to predict the distribution of survey responses.
This allows us not only to gain insight on the spectrum of beliefs held by NLP researchers, but also to uncover \textit{false sociological beliefs} where the community's predictions don't match reality.
We find such mismatches on a wide range of issues.
Among other results, the community greatly overestimates its own belief in the usefulness of benchmarks and the potential for scaling to solve real-world problems, while underestimating its own belief in the importance of linguistic structure, inductive bias, and interdisciplinary science.
\end{abstract}

\section{Introduction}
\label{sec:intro}

What do NLP researchers think about NLP?
\begin{itemize}
\item Are we devoting too many resources to scaling up?
\item Do language models understand language? Will they ever?
\item Is the traditional paradigm of model benchmarking still tenable?
\item What kinds of predictive models are ethical for researchers to build and release?
\item Will the next most influential advances come from industry or academic labs?
\end{itemize}

These questions and many more are actively debated in the research community, and views on them are a major factor in deciding what work gets done. Understanding the prevalence of different views on these issues is valuable for understanding the trajectory of NLP research and the structure of the field.
In addition, communication among researchers often rests on  \textit{sociological beliefs} about these questions: what people think people think. Getting these sociological beliefs wrong can slow down communication and lead to wasted effort, missed opportunities, and needless fights.
For example:
\begin{itemize}
    \item An early-career researcher may avoid researching a topic they think is important if they think that it is not valued by the community and it would be hard to get past review.
    \item Extra time and effort may be spent in papers and research agendas defending what is thought to be a provocative position, when it is in fact already well established and widely believed.
    \item Papers or public communications generally appeal to premises they believe are settled or accepted as general consensus, but if the premise is not actually settled, then the argument will not be convincing to a large portion of the community, and the discourse may become more fractured, with increasingly isolated groups of researchers forming echo chambers around different sets of assumptions.
\end{itemize}
The NLP research community gets to know itself in a variety of ways: discussions with colleagues at the same institution, presentations and interactions at conferences, invited talks and panel discussions, and interactions on social media like Twitter.
All of these sources of information are biased, for example through self-selection towards similar people and amplification of already-prominent or controversial voices.
These effects can make it difficult for any individual to get a sense of the NLP research community's beliefs as a whole.

For these reasons, we believe it is worth trying to objectively assess NLP researchers' collective stance on controversial issues.
So from May to June 2022, we conducted the \textbf{NLP Community Metasurvey}.
On each issue, we present a stance, such as \textit{Currently, the field focuses too much on scaling up machine learning models}~(\Q5-1), and ask respondents to state whether they agree or disagree.
Then we ask them to also \textit{predict the percentage of respondents who will agree}.
This gives us insight into the community's object-level beliefs as well as its sociological beliefs, and allows us to identify where the two may be misaligned.

This work is directly inspired by the PhilPapers Surveys~\citep{bourget2014what,bourget2020philosophers}, a research effort created and maintained by philosophers to assess the philosophy community's beliefs about current topics in their field, along with a separate metasurvey to assess philosophers' sociological beliefs about their professional community.
As \citeauthor{bourget-chalmers-conception} write:
\begin{displayquote}
Most of us have had the experience of reading philosophical papers that make sweeping sociological claims about the field that seem quite questionable. It is arguable that the ``received wisdom'' in a field (roughly, what people in a field take as a default view, one that can sometimes be presupposed) is not based on what most people in the field think, but rather is based on what most people think most people think. If the received wisdom is grounded in false sociological beliefs, that is worth knowing. It occurred to us that it would not be all that difficult to actually gather data about these matters, and that doing so might be an interesting and informative exercise.
\end{displayquote}

The rest of this document reports the methodology and results of the NLP Community Metasurvey.
Some key results include:
\begin{itemize}
    \item A large majority of respondents are against the hypothesis that scaling up current systems and methods could solve ``practically any important problem'' in NLP, and this view is perceived as much more popular than it is (\autoref{ssec:scale}, \Q2-1). Yet, narrow majorities of respondents also regard recent progress in large-scale modeling as progress towards AGI, and think AGI should concern NLP researchers (\autoref{ssec:agi}, \Q3-1, \Q3-2).
    \item A majority of respondents think that the scientific value of the majority of work in NLP is dubious (\autoref{ssec:state}, \Q1-5).
    \item A large majority of respondents believe that NLP researchers should give higher priority to incorporating insights from neighboring fields, and greatly underestimate the number of other NLP researchers that share this belief (\autoref{ssec:promise}, \Q5-7).
    \item Large majorities of respondents believe NLP research has had a positive impact on the world (\autoref{ssec:ethics}, \Q6-1), will have a positive impact in the future (\Q6-2), and could plausibly transform society (\autoref{ssec:agi}, \Q3-3). Despite this optimism, a substantial minority also foresee plausible risks of a major global catastrophe caused by ML systems (\Q3-4).
    \item A plurality of respondents think the most influential area of advances in the next 10 years will be in \textit{problem formulation and task design}, as opposed to \textit{hardware and data scaling}, which respondents believed would be the most popular opinion (\autoref{ssec:future}).
\end{itemize}
It is worth noting that our results are \textit{descriptive}, not prescriptive, as these issues cannot be resolved by majority vote.
By necessity, we are covering a subjectively chosen set of questions and reducing many complex issues into simplified scales, but we hope that these results can create common ground for fruitful discussion among the NLP research community.

\section{Methodology}
\label{label:design}

\paragraph{Choosing Questions}
We aimed to ask about issues:
\begin{itemize}
    \item which are frequently discussed in the community,
    \item which are the subject of public disagreement,
    \item about which the NLP community often reflects back on itself, especially where people seem to perceive themselves as in the minority (hot takes) or majority (taking something for granted), and
    \item for which, if we understand the community's opinions and meta-opinions better, it may aid our ability to communicate and help people understand how to most effectively communicate about their research.
\end{itemize}
With these criteria in mind,
we (the authors) brainstormed a large initial list of potential questions.
After discussing, we voted on which ones to include in the survey, chose roughly the top 30 questions, finalized the agree/disagree question format, and began pilot testing (described in \autoref{app:pilots}), which we used to refine the set of questions, their phrasing, and their presentation format in the survey. The questions used in the survey are shown in \autoref{fig:big-table-o-questions}.   

\paragraph{Target Demographic: Active Authors in ACL}
Since we are interested in the public and scientific discourse of the NLP community, we target the survey at active NLP researchers.
As an objective criterion, we define the target population of the survey as anyone who is an author on at least two *CL papers published in the last three years.
This definition allows us to assess response bias by comparing to ACL members~(\autoref{sec:demo}) and provides an objectively defined reference group for survey respondents to make predictions about in the meta-questions.

\paragraph{Question Format}
We present questions in thematic groupings (e.g., ``State of the Field''), phrased as statements that people could indicate their (dis)agreement with. 
They can respond to each statement on a 4-point scale of \Agree, \WeaklyAgree, \WeaklyDisagree, and \Disagree.
We choose not to include an option in the middle of the scale to indicate a neutral position because our intent is to push respondents to consider where they actually stand.
We instruct respondents to choose \WeaklyAgree\ or \WeaklyDisagree\ if they have even slight preferences for one side or the other (e.g., ``depends, leaning negative'').
However, it is sometimes the case that someone truly cannot make a judgment, and for these cases we include three \answerstyle{Other} answers:
\answerstyle{Question is ill-posed},
\answerstyle{Insufficiently informed on the issue},
and
\answerstyle{Prefer not to say}.

At the end of each section of the survey, we ask respondents to predict the proportion of respondents in our target population who will either \Agree\ or \WeaklyAgree\ with each statement. Respondents can answer with one of five buckets: 0--20\%, 20--40\%, 40--60\%, 60--80\%, or 80--100\%.
Respondents can also skip the meta-questions, but we encourage them to give their best guess even if they are not sure.
Finally, each section has a free-response box for the respondent to provide any comments, criticism, or other feedback on the survey.

The survey instructions are reproduced in full in \autoref{app:instructions}.

\paragraph{Platform and Distribution}

To host the survey, we used NYU Qualtrics.
Following the guidelines set out by the NYU Institutional Review Board (protocol FY2022-6461), all respondents gave informed consent before beginning the survey and could either refuse to answer (i.e., by responding \answerstyle{Prefer not to say}) or skip each question.

As an incentive for participation, we committed to donating \$10 for each respondent to one of several non-profit organizations that the respondent chooses at the end of the survey.\footnote{Our final donations were \$950 to the WHO COVID-19 Solidarity Response Fund (\url{https://www.who.int/emergencies/diseases/novel-coronavirus-2019/donate}), \$1,650 to GiveWell's Maximum Impact Fund (\url{https://www.givewell.org/maximum-impact-fund}), \$830 to GiveDirectly (\url{https://www.givedirectly.org/}), and \$1,140 to the Distributed AI Research Institute (\url{https://www.dair-institute.org/support}). 23 respondents (5\%) did not provide an answer to this question, so we did not make donations on their behalf.}

To attempt to reach a broad audience of NLP researchers,
we set up a homepage for the survey at \url{https://nlpsurvey.net}
and advertised in the following ways:
(a) ACL Member Portal: we sent a call for participation to the ACL membership mailing list. The email included the details of the survey, its purpose, and the charitable donation incentive. (b) ACL 2022 in Dublin: Four of our team members advertised the survey to conference attendees in-person. They distributed flyers/posters of our survey and free stickers that said ``NLP survey’’ or ``I took the NLP survey.’’ (c) Twitter: We released multiple tweets as advertisement, with the original being retweeted 100+ times. (d) Slack channels: We posted about the survey in the Slack channels of a few labs, as well as an NLP Slack channel with more than 470 members that was set up during ACL 2020. (e) Emails: We attempted to encourage more participants from senior authors by sending personal invitations to 568 authors that have published at least eight qualifying papers since 2019 (we did not exhaustively email all of them, as it required manually sourcing email addresses based on names in the ACL Anthology). (f) Other social media (including posting on WeChat to encourage participation from researchers in China) and personal interactions with NLP researchers. 

\section{Demographics}
\label{sec:demo}

\begin{figure}[t]
\begin{center}
\includegraphics[width=\textwidth]{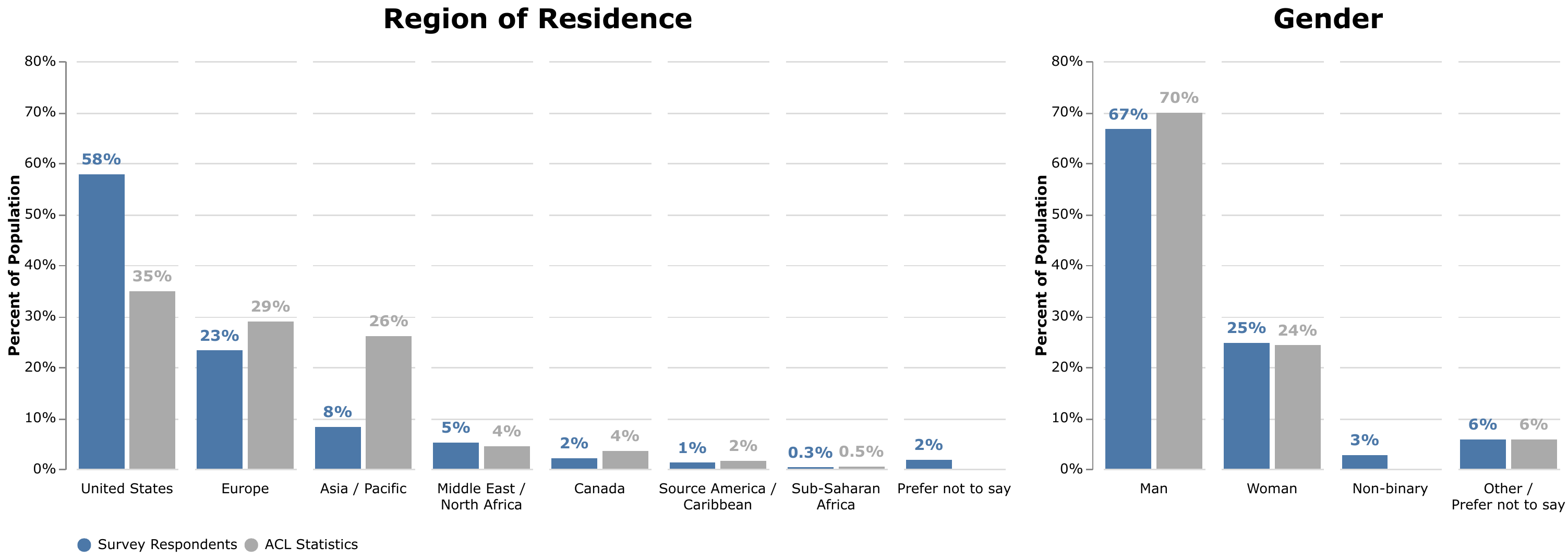}
\end{center}
\caption{Basic demographics of survey respondents, compared to available statistics from the ACL.
For region, we compare to ACL memberships as of summer 2021, and for gender, we use the diversity statistics currently available from the ACL, based on attendance at ACL 2017 in Vancouver (which lacked a specific ``non-binary'' category).}
\label{fig:acl-comparison}
\end{figure}

480 people completed the survey, of which 327 (68\%) are in our target demographic, reporting that they co-authored at least 2 ACL publications between 2019--2022.
We compute that 6323 people met this requirement during the survey period according to publication data in the ACL Anthology,
meaning we have survey responses from about 5\% of the total.
For the rest of this paper, we restrict all reported results to this subset.

Full question text and results for all demographics are available in \autoref{app:demographics}.

\subsection{Basic Demographics}
\autoref{fig:acl-comparison} shows location and gender statistics.
Survey respondents are mostly men (67\%) and mostly from the United States (58\%).
To get a sense of biases in our results, we compare to official statistics from the ACL.

\paragraph{Location}
The ACL publishes statistics about the countries of origin of its members.
While ACL members are not the same population as our target demographic, we can use them as a rough proxy to assess for geographic bias.
Comparing to the most recent available statistics\footnote{\url{https://www.aclweb.org/adminwiki/images/f/f4/Memberships_2021_by_Country_SUMMER.pdf}}
suggests that the United States is heavily overrepresented in our respondent population (58\% > 35\%), while Asia/Pacific is underrepresented (8\% < 26\%).
Asian countries with large ACL contingents were particularly underrepresented, including China (3\% < 9\%), India (1\% < 5\%), and Japan (0.8\% < 5\%).
We suspect this may largely be due to biases in our survey distribution methods (particularly Twitter and our personal networks).

\paragraph{Gender}
For gender, we compare to ACL-published diversity statistics, computed from the attendee population at ACL 2017 in Vancouver, Canada.\footnote{\url{https://www.aclweb.org/portal/content/acl-diversity-statistics}}
While our gender distribution is skewed, with 67\% men and 25\% women, it seems to roughly match the ACL population.

\paragraph{Underrepresented Minorities}
We ask respondents if they consider themselves to be part of an underrepresented minority group in NLP, to which 26\% report \textit{Yes}, 63\% report \textit{No}, and 11\% \textit{Prefer not to say}.

\subsection{Career}

\paragraph{Job Sector}
73\% of respondents are from academia, with 22\% from industry and 4\% in non-profit or government jobs.
Asked if their job is ``publication-oriented,''
89\% answer \textit{Yes}, including 62\% of respondents in industry, while 9\% answer \textit{No}.

\paragraph{Seniority}
Faculty and senior managers form a plurality of respondents at 41\%,
while 23\% are junior professionals (including postdocs), 33\% are PhD students, and 2\% are Masters students or undergraduates.
Breaking these down by sector, the largest group was academic PhD students (32\%),\footnote{The remaining 1\% of PhD students reported being in industry or declined to report their job sector.}
followed by faculty (29\%), senior managers (10\%), postdocs or other academic researchers (10\%), and junior researchers in industry (10\%).

Since ACL publishes statistics on the number of student memberships and regular memberships, we can also compare these numbers to the ACL population:
35\% of respondents are students, compared to 53\% of ACL members as of summer 2021.
While students are underrepresented in our results compared to ACL members as a whole, it is not clear whether they are underrepresented compared to our target demographic of recently published authors.

We also ask respondents to report the year they published their first research work in NLP.
20\% of respondents reported a year prior to 2009, with the earliest being 1985.
Another 20\% were in 2010--2014, 20\% in 2015--2017, 20\% in 2018--2019, and 10\% in 2019--2022, with the rest declining to answer.
See \autoref{app:demographics}, \autoref{fig:demog-firstpub} for more details.

\paragraph{Subfields} 
We ask respondents to report all subfields that fit their work from the last 3 years, providing a list derived from the tracks at recent *CL conferences (EMNLP 2021, ACL 2022, and NAACL 2022).
The most common responses are machine learning for NLP (39\%), interpretability and analysis (31\%), generation (28\%), resources and evaluation (27\%), and machine translation and multilinguality (26\%).
Coverage of the given subfields is broad, with each being marked by at least 20 respondents (6\%).
Full results are in \autoref{app:demographics}, \autoref{fig:demog-career}.

\paragraph{Research Activity} 
Respondents are highly active at ACL events, with 96\% saying they have attended an ACL event in the last 3 years.
Asked about their total number of peer-reviewed publications related to NLP,
13\% report 1--4,
47\% report 5--20, and
39\% report 21 or more.

\subsection{Other Information}

\begin{figure}[p]
\begin{center}
\tiny
      \begingroup
      \renewcommand*{\arraystretch}{1.4}
      \begin{tabulary}{\textwidth}{rL}
      \toprule
      \multicolumn{2}{c}{\textbf{1. State of the Field}} \\ \midrule 
\textbf{1-1.}
& \textbf{Private firms have too much influence.} Private firms have too much influence in guiding the trajectory of the field.\\
\textbf{1-2.}
& \textbf{Industry will produce the most widely-cited research.} The most widely-cited papers of the next 10 years are more likely to come out of industry than academia.\\
\textbf{1-3.}
& \textbf{NLP winter is coming (10 years).} I expect an "NLP winter" to come within the next 10 years, in which funding and job opportunities in NLP R\&D fall by at least 50\% from their peak.\\
\textbf{1-4.}
& \textbf{NLP winter is coming (30 years).} I expect an "NLP winter" to come within the next 30 years, in which funding and job opportunities in NLP R\&D fall by at least 50\% from their peak.\\
\textbf{1-5.}
& \textbf{Most of NLP is dubious science.} A majority of the research being published in NLP is of dubious scientific value.\\
\textbf{1-6.}
& \textbf{Author anonymity is worth it.} Author anonymity during review is valuable enough to warrant restrictions on the dissemination of research that is under review.\\
\midrule \multicolumn{2}{c}{\textbf{2. Scale, Inductive Bias, and Adjacent Fields}} \\ \midrule 
\textbf{2-1.}
& \textbf{Scaling solves practically any important problem.} Given resources (i.e., compute and data) that could come to exist this century, scaled-up implementations of established existing techniques will be sufficient to practically solve any important real-world problem or application in NLP.\\
\textbf{2-2.}
& \textbf{Linguistic structure is necessary.} Discrete general-purpose representations of language structure grounded in linguistic theory (involving, e.g., word sense, syntax, or semantic graphs) will be necessary to practically solve some important real-world problems or applications in NLP.\\
\textbf{2-3.}
& \textbf{Expert inductive biases are necessary.} Expert-designed strong inductive biases (à la universal grammar, symbolic systems, or cognitively-inspired computational primitives) will be necessary to practically solve some important real-world problems or applications in NLP.\\
\textbf{2-4.}
& \textbf{Ling/CogSci will contribute to the most-cited models.} It is likely that at least one of the five most-cited systems in 2030 will take clear inspiration from specific, non-trivial results from the last 50 years of research into linguistics or cognitive science.\\
\midrule \multicolumn{2}{c}{\textbf{3. AGI and Major Risks}} \\ \midrule 
\textbf{3-1.}
& \textbf{AGI is an important concern.} Understanding the potential development of artificial general intelligence (AGI) and the benefits/risks associated with it should be a significant priority for NLP researchers.\\
\textbf{3-2.}
& \textbf{Recent progress is moving us towards AGI.} Recent developments in large-scale ML modeling (such as in language modeling and reinforcement learning) are significant steps toward the development of AGI.\\
\textbf{3-3.}
& \textbf{AI could soon lead to revolutionary societal change.} In this century, labor automation caused by advances in AI/ML could plausibly lead to economic restructuring and societal changes on at least the scale of the Industrial Revolution.\\
\textbf{3-4.}
& \textbf{AI decisions could cause nuclear-level catastrophe.} It is plausible that decisions made by AI or machine learning systems could cause a catastrophe this century that is at least as bad as an all-out nuclear war.\\
\midrule \multicolumn{2}{c}{\textbf{4. Language Understanding}} \\ \midrule 
\textbf{4-1.}
& \textbf{LMs understand language.} Some generative model trained only on text, given enough data and computational resources, could understand natural language in some non-trivial sense.\\
\textbf{4-2.}
& \textbf{Multimodal models understand language.} Some multimodal generative model (e.g., one trained with access to images, sensor and actuator data, etc.), given enough data and computational resources, could understand natural language in some non-trivial sense.\\
\textbf{4-3.}
& \textbf{Text-only evaluation can measure language understanding.} We can, in principle, evaluate the degree to which a model understands natural language by tracking its performance on text-only classification or language generation benchmarks.\\
\midrule \multicolumn{2}{c}{\textbf{5. Promising Research Programs}} \\ \midrule 
\textbf{5-1.}
& \textbf{There's too much focus on scale.} Currently, the field focuses too much on scaling up machine learning models.\\
\textbf{5-2.}
& \textbf{There's too much focus on benchmarks.} Currently, the field focuses too much on optimizing performance on benchmarks.\\
\textbf{5-3.}
& \textbf{On the wrong track: model architectures.} The majority of research on model architectures published in the last 5 years is on the wrong track.\\
\textbf{5-4.}
& \textbf{On the wrong track: language generation.} The majority of research in open-ended language generation tasks published in the last 5 years is on the wrong track.\\
\textbf{5-5.}
& \textbf{On the wrong track: explainable models.} The majority of research in building explainable models published in the last 5 years is on the wrong track.\\
\textbf{5-6.}
& \textbf{On the wrong track: black-box interpretability.} The majority of research in interpreting black-box models published in the last 5 years is on the wrong track.\\
\textbf{5-7.}
& \textbf{We should do more to incorporate interdisciplinary insights.} Compared to the current state of affairs, NLP researchers should place greater priority on incorporating insights and methods from relevant domain sciences (e.g., sociolinguistics, cognitive science, human-computer interaction).\\
\midrule \multicolumn{2}{c}{\textbf{6. Ethics}} \\ \midrule 
\textbf{6-1.}
& \textbf{NLP's past net impact is good.} On net, NLP research has had a positive impact on the world.\\
\textbf{6-2.}
& \textbf{NLP's future net impact is good.} On net, NLP research continuing into the future will have a positive impact on the world.\\
\textbf{6-3.}
& \textbf{It is unethical to build easily-misusable systems.} It is unethical to build and publicly release a system which can easily be used in harmful ways.\\
\textbf{6-4.}
& \textbf{Ethical and scientific considerations can conflict.} In the context of NLP research, ethical considerations can sometimes be at odds with the progress of science.\\
\textbf{6-5.}
& \textbf{Ethical concerns mostly reduce to data quality and model accuracy.} The main ethical challenges posed by current ML systems can, in principle, be solved through improvements in data quality/coverage and model accuracy.\\
\textbf{6-6.}
& \textbf{It is unethical to predict psychological characteristics.} It is inherently unethical to develop ML systems for predicting people's internal psychological characteristics (e.g., emotions, gender identity, sexual orientation).\\
\textbf{6-7.}
& \textbf{Carbon footprint is a major concern.} The carbon footprint of training large models should be a major concern for NLP researchers.\\
\textbf{6-8.}
& \textbf{NLP should be regulated.} The development and deployment of NLP systems should be regulated by governments.\\

      \bottomrule
      \end{tabulary}
      \endgroup
\end{center}
\caption{All survey questions with agree/disagree answers. Only the full statements (not the bolded summaries) were shown to respondents. Besides these, the survey also asked for an opinion on likely sources of future advances in NLP (\autoref{ssec:future}) as well as respondent demographics (\autoref{sec:demo}, \autoref{app:demographics}).}
\label{fig:big-table-o-questions}
\end{figure}

\paragraph{Twitter Use}
A large majority of respondents use Twitter,
with 18\% saying \textit{I routinely post},
58\% saying \textit{I follow but don't often post},
and 21\% saying \textit{I don't have a Twitter account, rarely look at it, or don't use it for research or NLP related content}.
While we don't know what proportion of our target demographic uses Twitter,
it seems likely that this is one of the largest sources of bias in our respondent population.
At the same time, the purpose of this survey is partly to study NLP researchers' perceptions of the NLP community for the purpose of improving the public and scientific discourse.
To the extent that these perceptions are formed on Twitter, and the public discourse is carried out on Twitter, our results may be useful even if biased towards Twitter users.

\paragraph{Finding the Survey}
Asked how they heard about the survey,
37\% reported hearing about it through Twitter,
21\% from the email we sent through the ACL Member Portal,
19\% from another mailing list, a Slack channel, or other messaging forum,
15\% through word of mouth or personal communication, and
3\% through an advertisement at ACL 2022, where we put up flyers and passed out stickers.
Again, this suggests that our results are likely skewed towards Twitter users.

\paragraph{Meta-Question Confidence}
We ask respondents to report their confidence in their meta-question predictions.
5\% report being \textit{Completely confident},
32\% \textit{Fairly confident},
34\% \textit{Somewhat confident},
13\% \textit{Slightly confident},
13\% \textit{Not at all confident}, and
2\% \textit{Prefer not to say}.
Between these groups, the \textit{Completely confident}
had the highest mean absolute error on their meta-question predictions (20.9\%),
though this was only significantly ($p < 0.05$) different from the \textit{Slightly confident} group (18.5\%), which performed best, despite abstaining from predictions in fewer cases (5\%) than the completely confident group (11\%).
We considered removing meta-question answers from less confident respondents for the purpose of reporting results, but since there are few statistically significant differences and the less confident respondents actually performed better, we include all meta-question responses in the rest of this work for the sake of simplicity.

\section{Results}
\label{sec:results}

For reference,
all survey questions which took agree/disagree responses are shown in \autoref{fig:big-table-o-questions}.
In the rest of this section,
we will discuss the results of each section of the survey in detail.

\begin{figure}[t]
\begin{center}
\includegraphics[width=\textwidth]{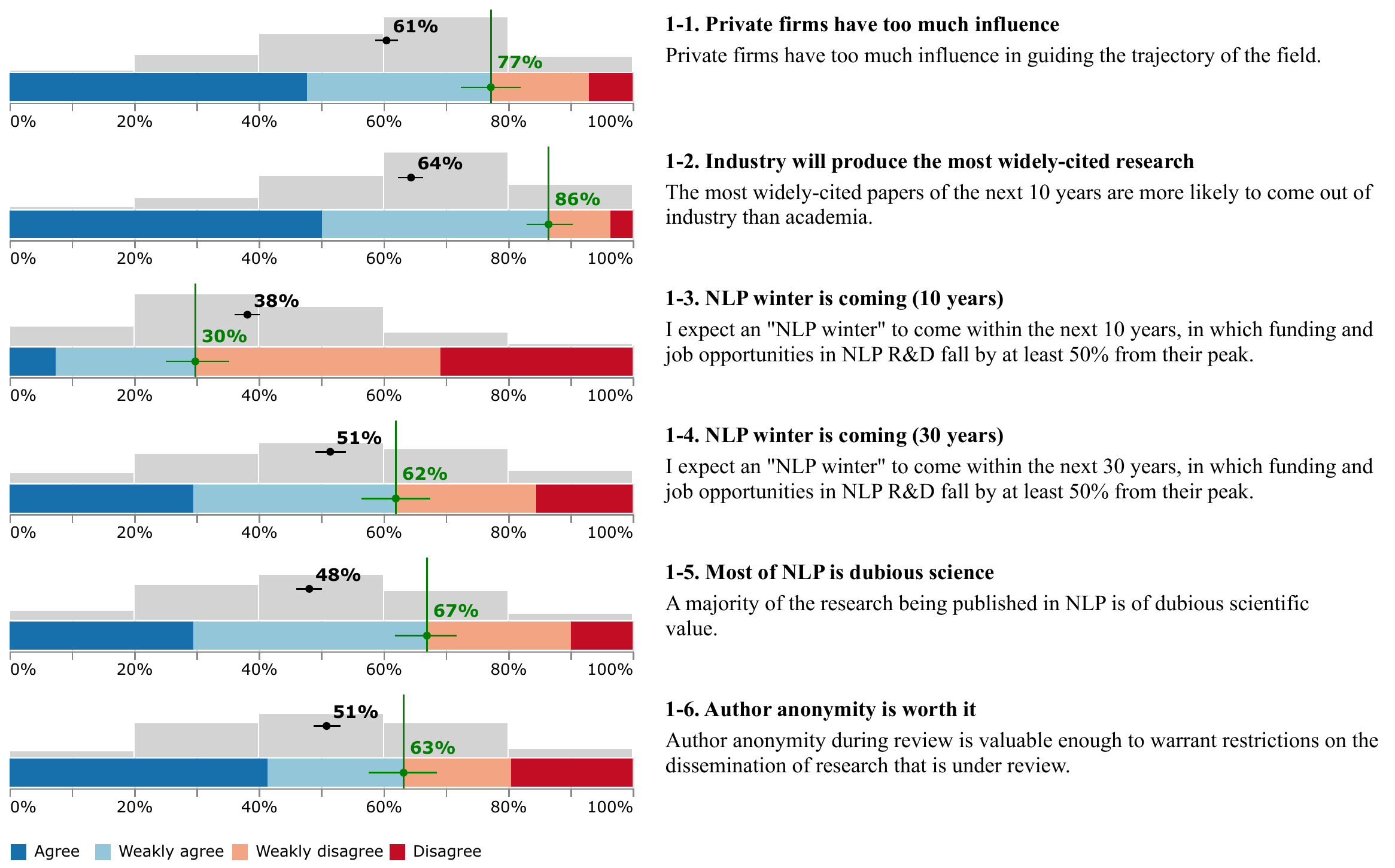}
\end{center}
\caption{\textit{State of the Field.}
Here, and in subsequent such figures, the lower number (in green) represents the fraction of respondents who agree with the position out of all those who took a side.
The grey bars show the relative proportion of meta-question predictions in each bin (0--20\%, 20--40\%, etc.), and the upper number (in black) shows the average predicted rate of agreement, computed treating each bin as its midpoint.
The green and black horizontal lines show 95\% bootstrap confidence intervals.
\label{fig:overview-state}}
\end{figure}

For each question (for example, in \autoref{fig:overview-state}), we display the proportion of \Agree, \WeaklyAgree, \WeaklyDisagree, and \Disagree\ answers in a band along the bottom of the visualization.
These percentages exclude those who gave one of the other answers (\answerstyle{Insufficiently informed on the issue}, \answerstyle{Question is ill-posed}, or \answerstyle{Prefer not to say}), which were relatively rare (<20\% of responses for all questions, <10\% for 75\% of questions; see \autoref{app:responses-overview}).
The vertical green line shows the total percentage who \Agree\ or \WeaklyAgree\ with the statement, which was the value we asked survey respondents to predict in the meta-questions.
The gray bars show the distribution of answers to the meta-questions, each bin aligned with its corresponding range of percentages (0\%--20\%, 20\%--40\%, etc.).
The green and black dots and bars show the mean and 95\% bootstrap confidence intervals of the true and predicted percentage of people who \Agree\ or \WeaklyAgree\ (treating each meta-question answer bucket as its midpoint).

Unless otherwise stated, all percentages and percentage differences mentioned in this section will be in absolute terms and exclude the `other' answers, and
when we refer to respondents ``agreeing'' with a statement (without special typesetting) we include both \Agree\ and \WeaklyAgree\ answers (and respectively with \Disagree\ and \WeaklyDisagree).
In this discussion we will sometimes break down results by demographic group (e.g., comparing the agreement rates of men and women); unless otherwise stated, all such comparisons correspond to statistically significant differences between the groups ($p$ < 0.05) by a bootstrap test.
More information, visualizations, and confidence intervals for such comparison can be found online at \url{https://nlpsurvey.net/results/}.

\subsection{State of the Field (\autoref{fig:overview-state})}
\label{ssec:state}

The first set of questions asks for opinions about the health of the NLP community.

\paragraph{Industry's Undue Influence (\Q1-1, \Q1-2)}
Private firms are overwhelmingly seen as likely to produce the most-cited research of the next 10 years (\Q1-2, 82\%), but they are also seen as having too much influence (\Q1-1, 74\%).
This suggests, as some respondents pointed out in the survey feedback, that many believe that number of citations is not a good proxy for value or importance.
It also suggests a belief that industry's continued dominance will have a negative effect on the field, perhaps through their singular control of foundational systems such as GPT-3~\citep{brown2020language} and PaLM~\citep{chowdhery2022palm}, or from the energy that widely-cited work in pretraining~\citep{devlin2019bert,radford2019language} draws away from other research agendas.
Respondents under-predict the popularity of the majority view by more than 15\% on both of these questions, suggesting they might believe alternative agendas are already under-prioritized, such as directions focusing on incorporating interdisciplinary insights as opposed to raw scaling, or problem formulation and task design---other under-predicted views, as we will see in \autoref{ssec:scale}, \autoref{ssec:promise}, and \autoref{ssec:future}).

The under-prediction of agreement on \Q1-1 and \Q1-2 may also be an artifact of our sample population, which is overwhelmingly academic.
Opinions are very different between job sectors, where 82\% in academia agree that private firms have too much influence (\Q1-1) compared to only 58\% of respondents in industry.

\paragraph{NLP Winter, Eventually (\Q1-3, \Q1-4)}

We ask respondents whether they expect there to be an ``NLP winter,'' where funding and job opportunities fall by at least 50\% from their peak, in the near future.
A substantial minority of 30\% expect this to happen within the next 10 years (\Q1-3), with only 7\% \Agree{}ing.
For the next 30 years (\Q1-4),
confidence is much greater, with 62\% expecting an NLP winter.
Even a minority predicting such a major shift in the field reflects an overall belief that NLP research will undergo substantial changes in the near future (at least, in who is funding it and how much).
Further interpretation of these results is difficult: For example, respondents may believe an NLP winter will arrive because the pace of innovation will stall (perhaps the reason they think industry research is overemphasized), because the ability to advance the state of the art will be monopolized by a small number of well-resourced industry labs (as they expect industry to continue producing widely-cited research), or because the distinction between NLP and other AI disciplines will disappear (as suggested by some respondents).

\paragraph{Dubious Science (\Q1-5)}
A majority agrees that most NLP work is of ``dubious scientific value'' (67\%).
Respondents expressed uncertainty over what should count as ``dubious,'' as well as concerns about who determines the value of research.
On one hand, such research could refer to work which is fundamentally unsound with ill-posed questions and meaningless results, which would be a powerful indictment of NLP research.
On the other hand, it could simply mean that many reported findings are of little importance or are not robust, which would arguably not make NLP unique among sciences~\citep{ioannidis2005why}. 
Either way, this result suggests that many NLP researchers think it is worth reflecting deeply on the value of our work. As respondents see the community being less critical than it actually is (by 19\% absolute), it might be that those who are critical of the scientific standards of the field are not as likely to voice their views in public, or that vocal critics who exist are seen as less representative of the population than they actually are.
 
\paragraph{Anonymity: Still Controversial (\Q1-6)}
*CL conferences have much stricter anonymity policies than many other conferences NLP researchers submit to (e.g., NeurIPS, ICLR, and ICML). Responses suggest the community is in favor of these policies on balance (63\% agree anonymity is important enough to warrant restrictions on disseminating preprints), though they are perceived as contentious: respondents guessed that around 51\% of the target population would be in favor of such restrictive policies. Since the *CL anonymity policies have been subject to intense debate on platforms such as Twitter, this suggests that those critical of the policies may have been disproportionately represented in the minds of NLP researchers.
This question was also split by gender, with 77\% of women agreeing but only 58\% of men---possibly due to concerns or experience with discrimination on the basis of author identity.

\begin{figure}[t]
\begin{center}
\includegraphics[width=\textwidth]{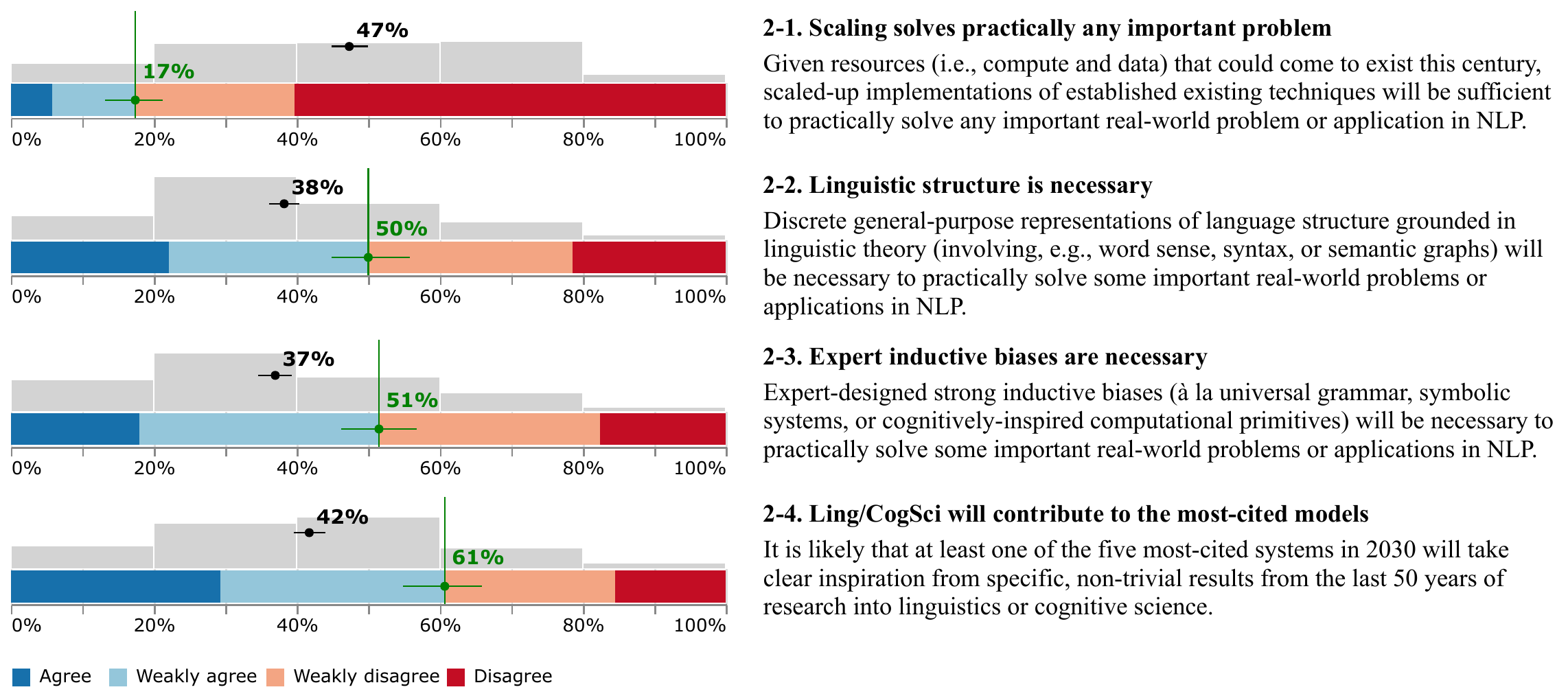}
\end{center}
\caption{\textit{Scale, Inductive Bias, and Adjacent Fields.}\label{fig:overview-scale}}
\end{figure}

\subsection{Scale, Inductive Bias, and Adjacent Fields (\autoref{fig:overview-scale})}
\label{ssec:scale}

Questions and meta-questions about the long-term potential of scale, inductive bias, and linguistic structure reveal some of the most striking mismatches between respondent attitudes and beliefs about those attitudes.
Broadly speaking, the pro-scale and anti-structure views were much less popular than respondents thought they would be. 

A common refrain in the era of ever-larger models is the \textit{Bitter Lesson}~\citep{sutton-2019-bitter}: ``General methods that leverage computation are ultimately the most effective, and by a large margin.''
Under this perspective,
one may expect benefits from incorporating linguistic structure or expert-designed inductive biases to be superseded by
learning mechanisms operating on fewer, more general principles if they have enough training data and model capacity.
While the success of deep learning and large language models may be taken as supporting evidence for the Bitter Lesson,
we find that the community has bought into the Lesson far less than it thinks it has.

\paragraph{Support for scaling maximalism is greatly overestimated (\Q2-1)}
We ask respondents for their views on a strong version of the Bitter Lesson: whether scaling up compute and data resources with established existing techniques would be sufficient to practically solve any important problem in NLP (\Q2-1).
Overall, this is seen as a controversial issue, with respondents predicting a roughly even split of 47\% agreement (though variance among predictions was high).
However, only a small minority (17\%) actually agree with the position, forming the largest discrepancy between predicted and actual opinions in the entire survey.
This suggests that the popular discourse around recent developments in scaling up~\citep{chowdhery2022palm} may not be reflective of the views of the NLP research community as a whole.

\paragraph{Trend reversals are predicted for linguistic theory and inductive bias (\Q2-2, \Q2-3, \Q2-4)}
The rest of the views articulated in this section were seen as less popular than \Q2-1, but in reality they were much more popular (albeit still controversial).
On what it will take to practically solve any important problem in NLP, 50\% agree that explicit linguistic structure will be necessary (\Q2-2), and 51\% say the same for expert-designed inductive biases (\Q2-3).
In addition, 61\% of respondents say it's likely that one of the five most-cited systems in 2030 will take inspiration from clear, non-trivial results from the last 50 years of linguistics or cognitive science research (\Q2-4).
All of these views are under-predicted by 12--19\%, though the predictions more closely match the responses given by men, as responses to these questions were split by gender.
Most notably, women are significantly more likely to agree with \Q2-2 that linguistic structure is necessary (65\%) compared to men (42\%). Women also agreed with \Q2-3 (62\%) and \Q2-4 (69\%) more than men (49\% and 57\%, respectively), but these differences were not statistically significant.

Like many respondents, we find these results surprising.
It seems that many believe there will be a reversal of the current trend of end-to-end modeling with low-bias neural network architectures.
The results for \Q2-4 are particularly surprising to us, as even \textit{today}'s most cited systems seem not to satisfy this requirement, building on little more from cognitive science than a rough construal of \textit{neurons}, \textit{attention}, and \textit{tokens}, which date back much further than 50 years.

\begin{figure}[t]
\begin{center}
\includegraphics[width=\textwidth]{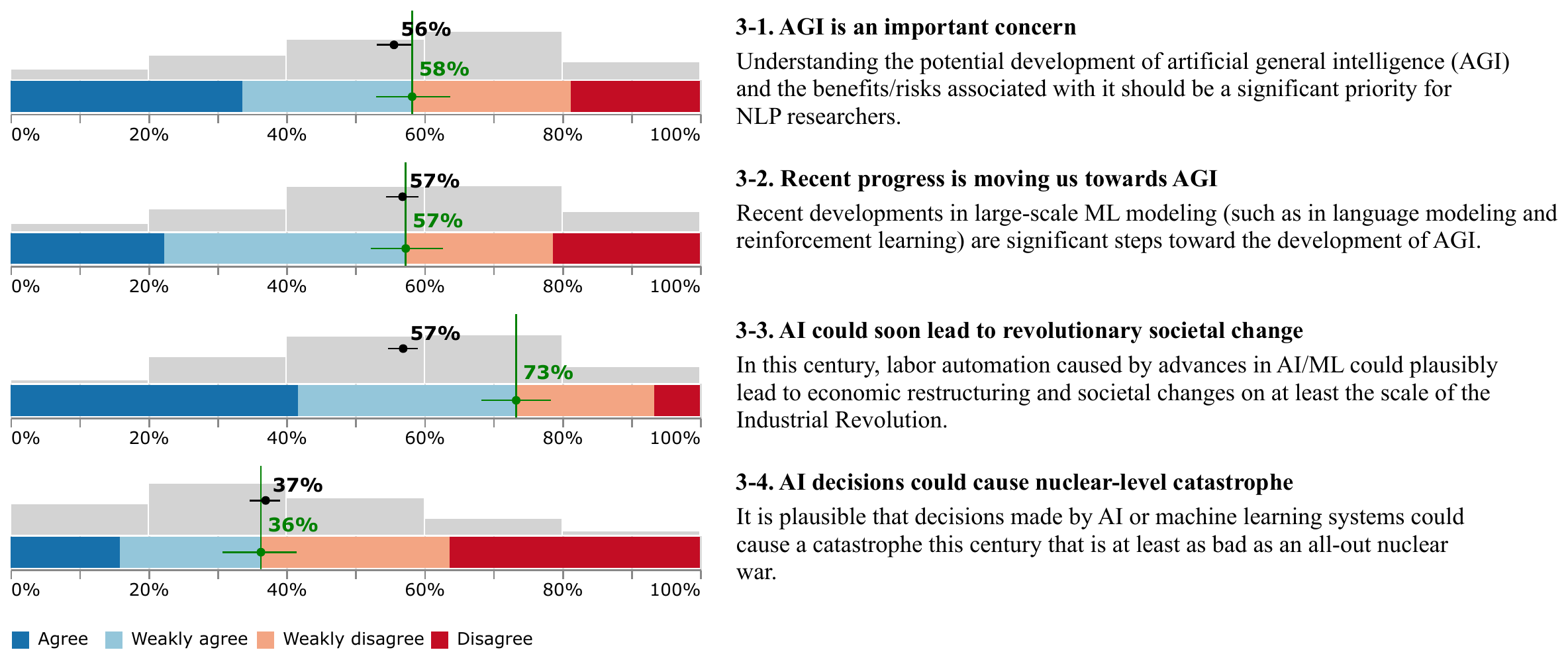}
\end{center}
\caption{\textit{Artificial general intelligence (AGI) and major risks}. \label{fig:overview-agi}}
\end{figure}

\subsection{AGI and Major Risks (\autoref{fig:overview-agi})}
\label{ssec:agi}

The versatility and impressive language output of large pretrained models such as GPT-3 \citep{brown2020language} and PaLM \citep{chowdhery2022palm} have
prompted renewed discussions about artificial general intelligence (AGI), including predictions of when it might arrive, whether we are actually advancing toward it, and what its consequences would be.
In this section, we ask about AGI and some of the largest possible impacts of AI technology.

One concern is that respondents' answers may depend on their definition of ``artificial general intelligence,'' and whether they think it is well-defined at all.
Our approach to this problem is to deliberately \textit{not} provide a definition (which some respondents would surely find objectionable, no matter which definition we choose).
Instead, we instruct respondents to answer according to their preferred definition, i.e., \textit{how they think the community should use the term}, as we view this as an important issue to assess when figuring out how to talk about the issue as a community.

\paragraph{AGI is a known controversy (\Q3-1, \Q3-2)}
On the questions explicitly about AGI, respondents were split near the middle,
with 58\% agreeing that AGI should be an important concern for NLP researchers (\Q3-1) and 57\% agreeing that recent research has advanced us toward AGI in some significant way (\Q3-2).
The two views are highly correlated,
with 74\% of those who think AGI is important also agreeing with \Q3-2 that we're progressing towards it, while only 37\% of people who don't think AGI is important think we're making that kind of progress.
The meta-responses split similarly to the object-level responses, indicating that the community has a good sense that this is a controversial issue.

It is worth acknowledging what this means:
AGI is a controversial issue,
the community in aggregate knows that it's a controversial issue,
and now (courtesy of this survey)
we can know that we know that it's controversial.
While some may believe that AGI is obviously coming soon,
and some may believe that it's obviously ill-defined,
taking either position for granted in the public discourse or scholarly literature
may not be an effective way to communicate to a broad NLP audience;
rather, careful and considered discussion of the issue will be more productive for building common ground.

\paragraph{Revolutionary and catastrophic outcomes are a concern (\Q3-3, \Q3-4)}
73\% of respondents agree that labor automation from AI could plausibly lead to revolutionary societal change in this century, on at least the scale of the Industrial Revolution (\Q3-3).
This points to a common reason why those who agree with \Q3-1 might think AGI is an important concern, especially if we are meaningfully progressing towards it (\Q3-2), as it could be fundamentally transformative for society;
indeed, all views expressed in this section are positively correlated (see \autoref{sec:correlation}).
But it's worth noting that a significant fraction of respondents (23\%) agree with the prospect of revolutionary change (\Q3-3) while disagreeing with the importance of AGI,
suggesting that discussions about long-term or large-scale impacts of our work in NLP may not need to be tied up in the AGI debate.

About a third (36\%) of respondents agree that it is plausible that AI could produce catastrophic outcomes in this century, on the level of all-out nuclear war (\Q3-4).
While this is a much smaller proportion than those who expect revolutionary societal change (\Q3-3), the stakes are extremely high and a substantial minority expressing concern about such outcomes indicates that a deeper discussion of such risks may be warranted in the NLP community.
While we do not ask about specific ways in which respondents think this could happen, potential reasons for such concerns are discussed by \citet{bostrom2014superintelligence,amodei2016concrete} and \citet{hubinger2019risks}.
Certain demographics, particularly women (46\%) and underrepresented minority groups (53\%), were more likely to agree with \Q3-4,
reflecting pessimism about our ability to manage dangerous future technology perhaps based in the present-day track record of disproportionate harms to these groups.

\Q3-4 received a lot of critical feedback.
Some respondents object to ``all-out nuclear war'' as far too strong, saying they would agree with less extreme phrasings of the question.
This suggests that our result of 36\% is an underestimate of respondents who are seriously concerned about negative impacts of AI systems.
Some respondents also comment that AI/ML systems should not be discussed as if they have agency to make decisions,
as all AI ``decisions'' can be traced back to human decisions regarding training data, architecture, how and on what phenomena models are evaluated (or not), and deployment decisions, among other factors.

\begin{figure}[t]
\begin{center}
\includegraphics[width=\textwidth]{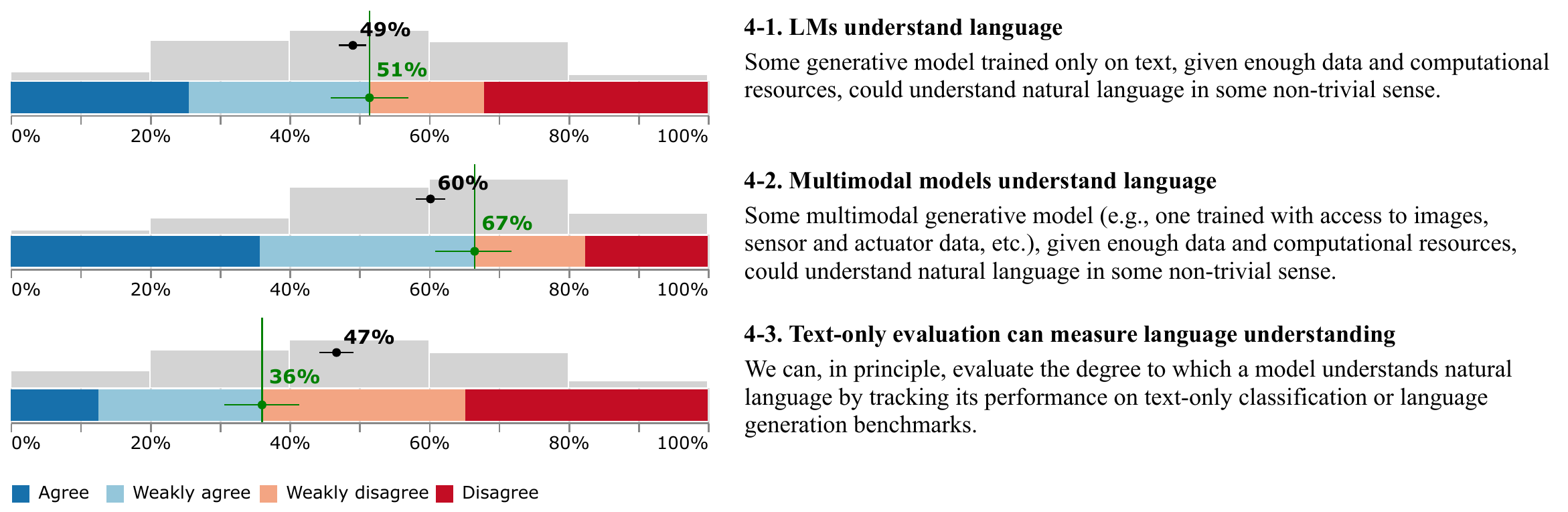}
\end{center}
\caption{\textit{Language Understanding}.\label{fig:overview-understanding}}
\end{figure}

\subsection{Language Understanding (\autoref{fig:overview-understanding})}
\label{ssec:understanding}

The question of whether language models understand language has been the subject of some debate in the community~\citep[\S{}2.6]{bender-koller-2020-climbing,merrill-etal-2021-provable,bommasani2021foundation}.
In this section, we ask some questions relevant to the issue,
but one of the challenges is that their answers are highly dependent on how one defines the word ``understand.''
For this reason, as with \autoref{ssec:agi}, we deliberately choose \textit{not} to provide a definition, as doing so would risk begging the question or forcing a definition that some would certainly find objectionable.
Instead, we instruct respondents to answer according to their preferred definitions, i.e., \textit{how they think the community should use} the word ``understand,'' as we view this as an important element of the discussion.
Many respondents commented that this choice made it harder to respond to the questions in this section, and said they would have preferred a set definition,
but only 3--5\% responded to any of these questions with \answerstyle{Question is ill-posed}.

\paragraph{LMs understanding language is a known controversy (\Q4-1, \Q4-2)}
The question of whether language models can understand language (\Q3-1) was split right down the middle, with 51\% agreeing.
This controversy is reflected in people's predictions as well,
which average to an estimate of 49\% agreement.
Many more people (67\%) agree once the model has access to multimodal data (images, etc.).
As with the importance of AGI (\autoref{ssec:agi}),
whether language models understand language is known to be controversial, and the results of this survey can make it known that it is known. 
So whatever one's views are on the issue,
it will likely be less useful to take those views for granted as a premise
when communicating to a broad NLP audience in the public discourse or scholarly literature.
Again, careful and considered discussion of the issue will likely be more productive for building common ground.

\paragraph{Understanding may be learnable, but not measurable, using text (\Q4-3)}
On the question of whether text-only evaluations can measure language understanding~(\Q4-3), the distribution of predictions was similar to that for language understanding by LMs (\Q4-1), averaging 47\% predicted agreement.
However, unlike \Q4-1, only 36\% actually agreed with the statement, suggesting that many view it as a separate issue, and that some may believe that there are things which are learnable from text alone, but cannot be measured using text alone.

Responses to questions in this section vary considerably with respondents' gender and location.
On LMs understanding language (\Q4-1), men are more likely to agree (58\%) than women (37\%), and people in the US are more likely to agree (61\%) than those in Europe (31\%).
There is also a significant gender difference on \Q4-3 regarding text-only evaluation of language understanding, where 43\% of men agree as opposed to 21\% of women.

\begin{figure}[t]
\begin{center}
\includegraphics[width=\textwidth]{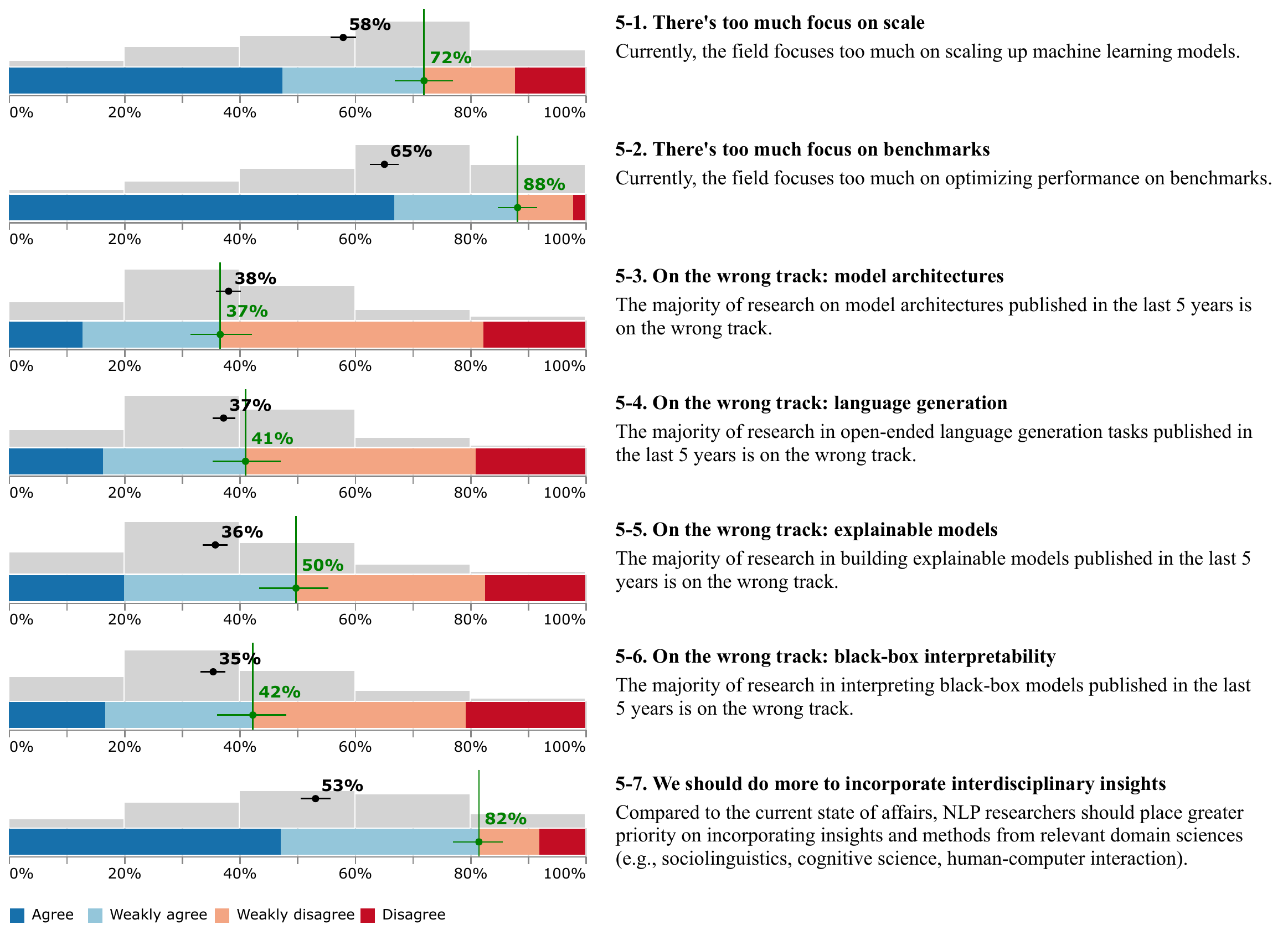}
\end{center}
\caption{\textit{Promising Research Programs}.\label{fig:overview-promise}}
\end{figure}

\subsection{Promising Research Programs (\autoref{fig:overview-promise})}
\label{ssec:promise}

In this section, we ask respondents about the kind of research they think the community should be doing, and which research directions they believe are not heading in the right direction.
We choose research agendas to ask about based on criticisms, debates, or findings in the literature and public sphere,
for example regarding current practice in benchmarking~\citep{bowman-dahl-2021-will,raji2021ai},
the relative value of advances in model architectures~\citep{narang-etal-2021-transformer,tay2022scaling},
the use of language models for generation tasks~\citep{bender2021dangers},
and explainability and interpretability of black-box models~\citep{feng-etal-2018-pathologies,jain-wallace-2019-attention,wiegreffe-pinter-2019-attention}.

\paragraph{Scaling and benchmarking are seen as over-prioritized (\Q5-1, \Q5-2)} Over 72\% of respondents believe that the field focuses too much on scale (\Q5-1), a view that was underestimated at 58\%.
This reflects the same pattern as \Q2-1, where the prevalence of pro-scale views is overestimated.
An even stronger majority of 88\% believe there is too much focus on optimizing performance on benchmarks (\Q5-2), a view that is highly correlated with \Q5-1 (see \autoref{sec:correlation}) and is similarly under-predicted at 65\%.

\paragraph{On the wrong track? Opinions vary (\Q5-\{3--6\})}
We ask whether four specific research directions are ``on the wrong track'': model architectures (\Q5-3), open-ended generation tasks (\Q5-4), explainable models (\Q5-5), and black-box interpretability (\Q5-6).
Respondents are divided on these questions, with agreement rates between 37\% and 50\%, reflecting that these are controversial issues.
In most cases, respondents' predictions also reflect this divide, with a possible exception in explainability (\Q5-5), where 50\% true agreement is under-predicted at 36\%, reflecting that more community members are critical of research in explainable modeling than expected.

While we deliberately used the vague phrase ``on the wrong track'' to get a sense of people's general attitudes, 
some respondents took issue with the framing of these questions; for example, one asks if it means \textit{asking the wrong question} or \textit{finding the wrong solutions}.
As such, respondents' precise interpretations of these questions may vary.

\paragraph{Interdisciplinary insights are valued more than we think (\Q5-7)}
The largest disparity between predicted and actual results in this section is on \Q5-7, stating that NLP researchers should do more to incorporate insights from relevant domain sciences.
While respondents' predictions about the community's opinions split this issue down the middle (53\%), in reality 82\% agree with the view (an outcome only expected by 11\% of respondents).

This raises a question:
If so many people agree that we should place greater priority on interdisciplinary work (\Q5-7),
why isn't more such work already happening?
One possible explanation is that the responses to \Q5-7 are a form of wishful thinking: Few believe that scale will be sufficient to solve our problems (\Q2-1, \Q5-1), and many think benchmarks are overemphasized (\Q5-2) and insights from sciences like linguistics and cognitive science will be necessary for long-term progress (\Q2-2, \Q2-3).
However, perhaps few know how to actually get results or useful insights from an interdisciplinary approach,
leading this kind of work to be underrepresented in the literature and public discourse despite high demand for it.
This suggests that the real issue may not be that NLP researchers do not assume interdisciplinary work has anything to offer so much as that we lack the knowledge and tools to make such work effective.

One caveat with this result is that responses vary significantly by job sector; 85\% of those in academia agree with \Q5-7 compared to 68\% of those in industry, and our survey is mostly academics.
Despite this difference, even the industry-only agreement rate is underpredicted, so survey response bias likely does not fully explain the mismatch.

\begin{figure}[t]
\begin{center}
\includegraphics[width=\textwidth]{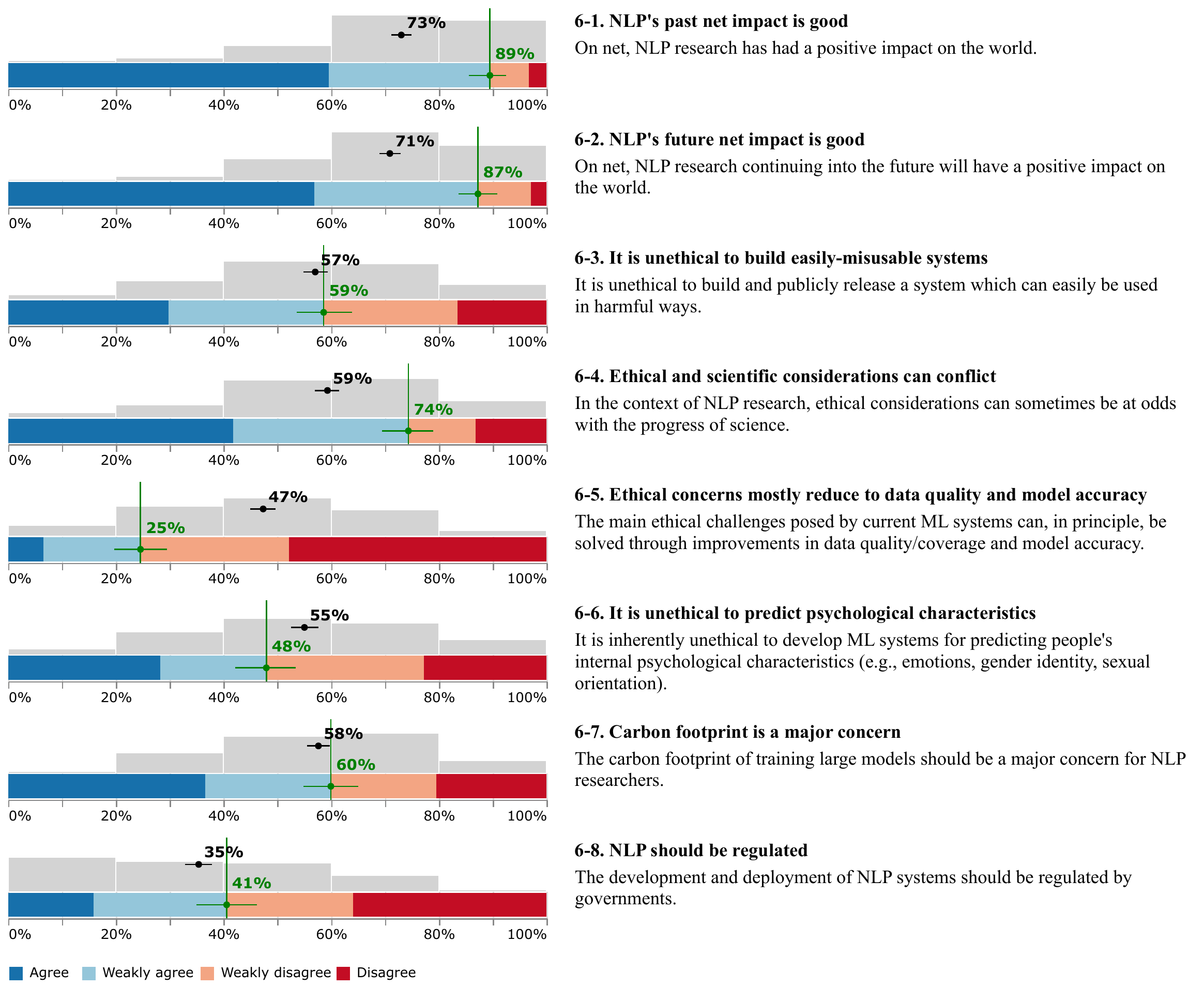}
\end{center}
\caption{\textit{Ethics}.\label{fig:overview-ethics}}
\end{figure}

\subsection{Ethics (\autoref{fig:overview-ethics})}
\label{ssec:ethics}

\paragraph{NLP is seen as good, and maybe extremely good (\Q6-1, \Q6-2)}
Respondents overwhelmingly regard NLP as having a positive overall impact on the world, both up to the present day (89\%, \Q6-1) and going into the future (87\%, \Q6-2).
This strong endorsement of NLP's future impact stands in contrast with more substantial worries about catastrophic outcomes (36\%, \Q3-4).
While the views are anticorrelated,\footnote{32\% of respondents who agreed that NLP will have a positive future impact on society (\Q6-2) also agreed that there is a plausible risk of catastrophe (\Q3-4), compared to 60\% reporting a belief in plausible catastrophic risk among those who disagreed with \Q6-2.}
a substantial minority of 23\% of respondents agreed with both \Q6-2 and \Q3-4, suggesting that they may believe NLP's potential for positive impact is so great that it even outweighs plausible threats to civilization.  
Whatever this means, it seems clear that many researchers think the stakes will be high in the near future when it comes to the impact of NLP research.

Interestingly, agreement with \Q6-1 and \Q6-2 are both underpredicted by more than 15\%, suggesting that pessimistic voices may be overrepresented in the public discourse.

\paragraph{Responsibility for misuse: Researchers are somewhat split (\Q6-3)}
In \Q6-3, we ask respondents if they think ``it is unethical to build and publicly release a system which can easily be used in harmful ways.''
This is admittedly vague, and its answer depends on many factors (e.g., how ``easily'' the system can be used, how it is released, etc.).
Our intent with the question is to get a sense of the degree to which respondents feel that researchers bear ethical responsibility for downstream misuse of the systems that they produce, and assess whether the community's views of itself are accurate in these terms.
Responses are somewhat split, with a majority of 59\% agreeing, and respondent predictions were reasonably accurate, averaging at 57\% predicted agreement.
But responses varied by gender, with 74\% of women agreeing versus 53\% of men.
It is worth comparing \Q6-3 to Article 1.1 of the ACM Code of Ethics,\footnote{\url{https://www.acm.org/code-of-ethics}}
which is adopted by the ACL\footnote{\url{https://www.aclweb.org/portal/content/acl-code-ethics}}
and states (among other things): ``Computing professionals should consider whether the results of their efforts will... be used in socially responsible ways.''

\paragraph{Belief in ethical/scientific conflict is underestimated (\Q6-4)}
When asked if ethical considerations can sometimes be at odds with scientific progress, 74\% of respondents agreed---considerably more than the average predicted agreement rate of 59\%.

There are a couple of potential interpretations of disagreement with \Q6-4.
On one hand, respondents may believe any ethical problems that come up during the course of NLP research can be solved easily or are trumped by the benefits of scientific progress.
On the other hand, they might believe that scientific `progress' which is ethically regressive should not count as `progress' or is inevitably pseudoscientific.
Several views in line with the latter (and none with the former) were expressed in the survey feedback, suggesting that it is likely the dominant interpretation among those who disagree with \Q6-4.
As disagreement was significantly overpredicted by survey respondents, this view may be overrepresented in the public sphere relative to the proportion of NLP researchers who hold it.

\paragraph{Reduction of ethics to data/accuracy is overestimated (\Q6-5)}
In light of public debates about the sources and nature of the harms caused by machine learning systems~\citep{kurenkov2020lessons}, we ask whether the main ethical challenges posed by current ML systems can be reduced to issues with data quality and model accuracy (\Q6-5).
It is estimated to be a common view, averaging 47\% predicted agreement, but is actually fairly uncommon, with only 25\% of respondents agreeing.

\paragraph{Predicting psychological characteristics is controversial, with caveats (\Q6-6)}
In light of discussions about surveillance and digital physiognomy~\citep{aguerayarcas2017physiognomy}, we ask whether it is inherently unethical to develop ML systems for predicting internal psychological characteristics like emotions, gender identity, and sexual orientation. Responses were split, with 48\% agreeing.

This question received a lot of critical feedback, and
it is unclear how much of this split is due to differences in opinion versus interpretations of the question.
Some respondents object to grouping transient states (e.g., emotion) with persistent traits (gender identity, sexual orientation), or say their answer depends on whether the trait is legally protected.
Some say it depends on the inputs available to the model, and others say that it may not be \textit{inherently} unethical but is ethically permissible in only a tiny set of carefully considered use cases.
Which of these elements of context respondents assumed may have played a major role in determining their answers, and future surveys on these issues might benefit from splitting \Q6-6 into several different questions.

\paragraph{Carbon footprint is a concern for many (\Q6-7)}
A majority of 60\% agree with the statement that the carbon footprint of training large models should be a major concern for NLP researchers (\Q6-8).
This concern is based in part on trends in computation for machine learning at large scale, as \citet{schwartz2020green} note a 300,000x increase in computation over 6 years leading up to 2019.
Following this, \citet{patterson2022carbon} argue that advances in model efficiency and energy management can soon lead to a plateau in energy use from training machine learning models.
Both argue that accountability and reporting of energy use is important for keeping the future carbon footprint of training ML models under control.
The responses to \Q6-8 indicate that a majority of the community would likely appreciate explicit reporting of energy use in NLP publications as well as work that increases the compute efficiency of model training.
Responses to this question varied greatly by gender, with 78\% of women agreeing as opposed to 51\% of men.

\paragraph{NLP researchers are skeptical of regulation (\Q6-8)}
Finally, we ask if the development and deployment of NLP systems should be regulated by governments (\Q6-8).
41\% of respondents agree, and while respondents' predictions are accurate on average, a large contingent (31\%) of respondents predicted a very low agreement rate of 0--20\%.
We intend \Q6-8 as a weak statement, i.e., that there should be \textit{any} regulations around the development and deployment of NLP systems.
However, respondents ask for more nuance, remarking that the answer depends on development versus deployment, details about use cases, and whether we only mean NLP-specific regulations or also include more general regulations on things like energy use or data privacy.
As respondents may have come to this question with different assumptions or interpretations around such issues, it is hard to read into the specific implications of this result,
except that respondents express a general skepticism of government regulation.

\begin{figure}[t]
\begin{center}
\includegraphics[width=\textwidth]{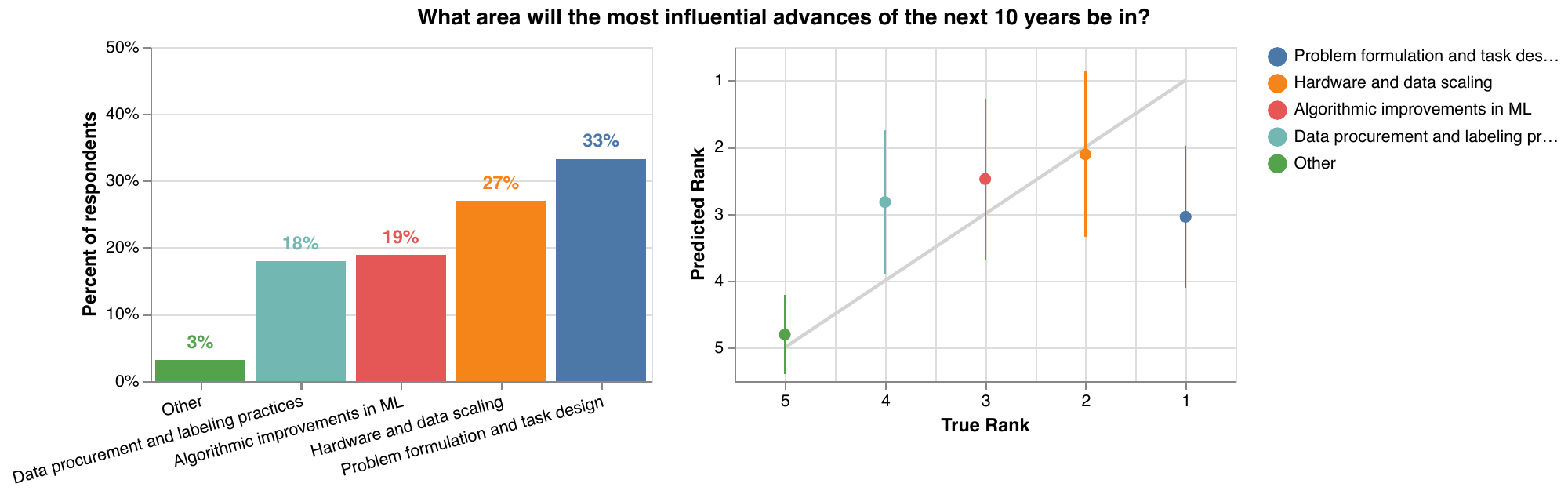}
\end{center}
\caption{%
\textit{Likely Sources of Future Advances}.
The left shows the distribution of answers, and the right shows predicted ranks (mean and standard deviation) of each answer relative to its true rank.
\label{fig:overview-advances}}
\end{figure}

\subsection{Likely Sources of Future Advances (\autoref{fig:overview-advances})}
\label{ssec:future}
In addition to the agree/disagree questions that constitute most of the survey,
we also ask respondents where they think the most influential advances of the next 10 years will come from, providing four choices: \textit{Hardware and data scaling}, \textit{Algorithmic improvements in ML}, \textit{Data procurement and labeling practices}, and \textit{Problem formulation and task design}. We also provide an ``other'' option for people to specify their own answer.
As the meta-question, we ask respondents to rank the answers from most popular (1) to least popular (5).\footnote{20\% of respondents rank ``Other'' on top (rank 1) for this question, even though very few actually provided it as their answer. It seems to us that these respondents probably ranked the answers backwards by mistake. To correct for this, we reverse the rankings provided by everyone who ranked ``Other'' first. While not a surefire fix, it doesn't seem to change any major trends in the results (\autoref{app:cleaning}, \autoref{fig:advances-adjustment}).}

The results reveal one surprise:
the popularity of the top answer, \textit{Problem formulation and task design}, was greatly underestimated.
A plurality of 33\% of respondents gave this answer, but it was only ranked first by 12\% of people and it ranked third on average in respondents' predictions.
Besides this, predictions roughly tracked reality (although noisy).
This suggests that there is a fairly common but underappreciated belief in the NLP community that researchers should be working on new ways of formulating the problems we're trying to solve, and that such work could have high impact.

\section{Correlation Analysis}
\label{sec:correlation}

Having each survey respondent provide opinions on a wide range of issues gives us the opportunity to examine the relationships between these opinions: Which sets of beliefs often come together, and which don't?
Also, are there demographic characteristics that correlate with certain beliefs?
To get a sense of this, we compute pairwise Spearman (rank-order) correlations between pairs of questions~(\autoref{ssec:correlations-question}) and demographic characteristics~(\autoref{ssec:correlations-demographic}),
and perform a clustering analysis on our results using PCA~(\autoref{ssec:clustering}).

\subsection{Question Correlations}
\label{ssec:correlations-question}

\begin{figure}
    \centering
    \includegraphics[width=\linewidth]{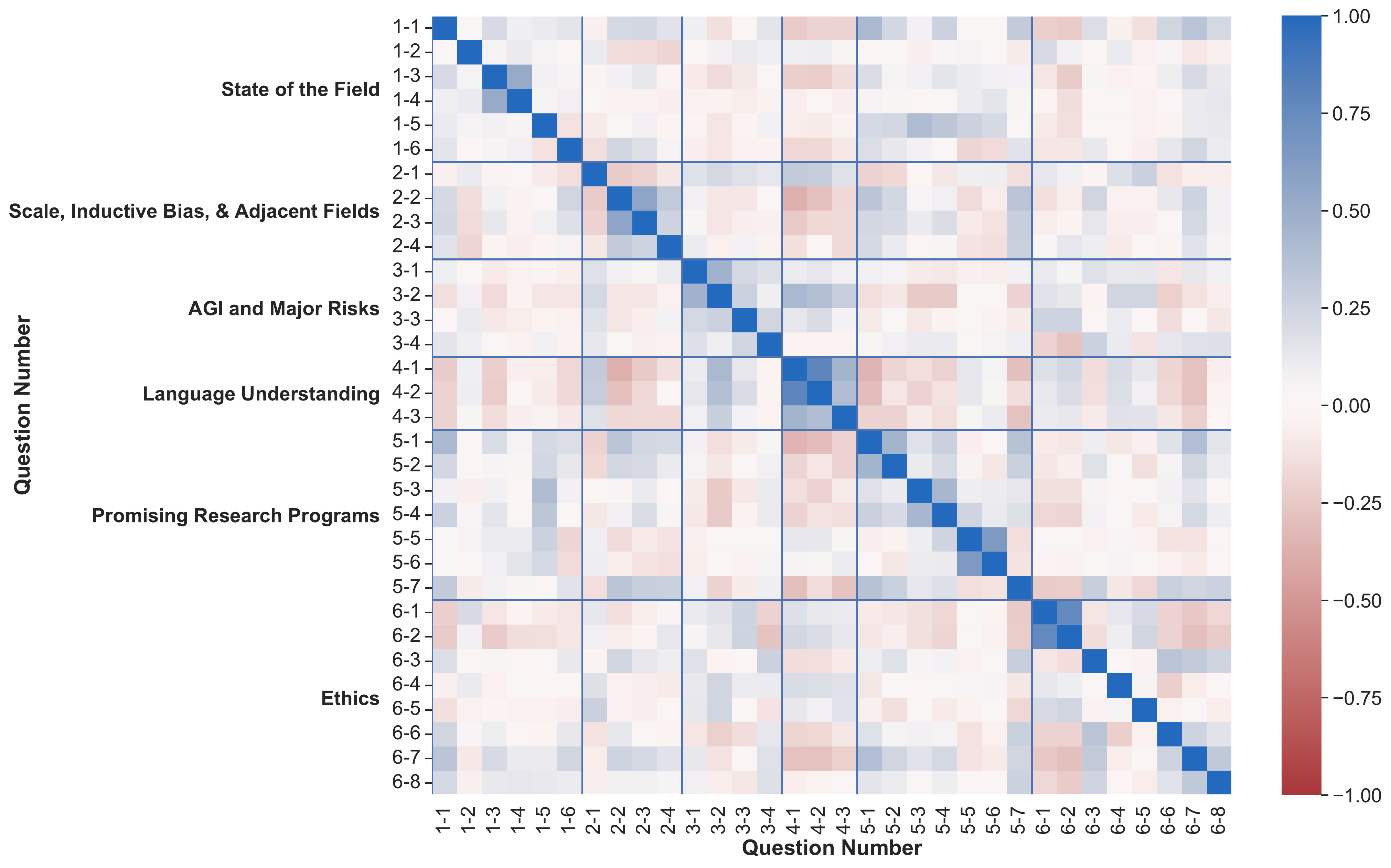}
    \caption{Spearman correlations between answers to all pairs of agree/disagree questions. Lines separate sections of the survey. Question numbers are given in \autoref{fig:big-table-o-questions}.}
    \label{fig:q_to_q_spearman}
\end{figure}

Spearman (rank-order) correlations between agree/disagree questions are shown in \autoref{fig:q_to_q_spearman}.
To compute these,
we order the response values as [\Disagree, \WeaklyDisagree, \answerstyle{Other}, \WeaklyAgree, \Agree],
where \answerstyle{Other} includes \answerstyle{Insufficiently informed on the issue}, \answerstyle{Question is ill-posed}, and \answerstyle{Prefer not to say}.\footnote{Even though the \answerstyle{Other} answers may not be meaningfully ``in between'' agreeing and disagreeing,
we find that running the analysis excluding \answerstyle{Other} options produces nearly identical results, so we keep them in order to apply the analysis to all of the data.}
Unsurprisingly, correlations tend to be stronger within a single section of the survey --- for example, perspectives on linguistic structure and inductive bias~(\autoref{ssec:scale}), AGI~(\autoref{ssec:agi}), language understanding~(\autoref{ssec:understanding}), and NLP's net impact on society (\autoref{ssec:ethics}, \Q6-1, \Q6-2).
Beyond these,
\autoref{tab:top_q_pairs} shows the highest-magnitude correlations between questions in different survey sections.

\begin{table}[h]
    \centering
    \resizebox{\linewidth}{!}{
    \begin{tabular}{llr}
    \toprule
    Q1 & Q2 & $\rho_s$ \\
    \midrule
\textbf{\Q1-1.} Private firms have too much influence. & \textbf{\Q5-1.} There's too much focus on scale. & \affirm{+0.43} \\
\textbf{\Q3-2.} Recent progress is moving us towards AGI. & \textbf{\Q4-1.} LMs understand language. & \affirm{+0.42} \\
\textbf{\Q1-5.} Most of NLP is dubious science. & \textbf{\Q5-3.} On the wrong track: model architectures. & \affirm{+0.40} \\
\textbf{\Q5-1.} There's too much focus on scale. & \textbf{\Q6-7.} Carbon footprint is a major concern. & \affirm{+0.38} \\
\textbf{\Q1-5.} Most of NLP is dubious science. & \textbf{\Q4-2.} Multimodal models understand language. & \affirm{+0.38} \\[0.4em]
\textbf{\Q2-2.} Linguistic structure is necessary. & \textbf{\Q4-1.} LMs understand language. & \negate{-0.38} \\
\textbf{\Q4-1.} LMs understand language. & \textbf{\Q5-1.} There's too much focus on scale. & \negate{-0.35} \\
\textbf{\Q1-1.} Private firms have too much influence. & \textbf{\Q6-7.} Carbon footprint is a major concern. & \affirm{+0.35} \\
\textbf{\Q2-2.} Linguistic structure is necessary. & \textbf{\Q5-7.} We should incorporate more interdisciplinary insights. & \affirm{+0.35} \\
\textbf{\Q2-2.} Linguistic structure is necessary. & \textbf{\Q5-1} There's too much focus on scale.  & \affirm{+0.34} \\[0.4em]
\textbf{\Q1-5.} Most of NLP is dubious science. & \textbf{\Q5-4.} On the wrong track: language generation. & \affirm{+0.33} \\
\textbf{\Q1-1.} Private firms have too much influence. & \textbf{\Q5-7.} We should incorporate more interdisciplinary insights. & \affirm{+0.30} \\
\textbf{\Q4-2.} Multimodal models understand language. & \textbf{\Q5-1.} There's too much focus on scale. & \negate{-0.30} \\
\textbf{\Q4-1.} LMs understand language. & \textbf{\Q5-7.} We should incorporate more interdisciplinary insights. & \negate{-0.30} \\
\textbf{\Q2-1.} Scaling solves practically any important problem. & \textbf{\Q4-1.} LMs understand language. & \affirm{+0.30} \\[0.4em]
\textbf{\Q2-2.} Linguistic structure is necessary. & \textbf{\Q4-2.} Multimodal models understand language. & \negate{-0.30} \\
\textbf{\Q2-1.} Scaling solves practically any important problem. & \textbf{\Q4-2.} Multimodal models understand language. & \affirm{+0.29} \\
\textbf{\Q4-2.} Multimodal models understand language. & \textbf{\Q6-7.} Carbon footprint is a major concern. & \negate{-0.28} \\
\textbf{\Q3-2.} Recent progress is moving us towards AGI. & \textbf{\Q4-3.} Text-only evaluation can measure language understanding. & \affirm{+0.28} \\
\textbf{\Q5-7.} We should incorporate more interdisciplinary insights. & \textbf{\Q6-3.} It is unethical to build easily misusable systems. & \affirm{+0.28} \\
    \bottomrule
    \end{tabular}}
    \caption{Top 20 Spearman correlations ($\rho_s$) between pairs of questions from distinct categories (i.e., with distinct first numbers). Correlations |$\rho_s| > 0.11$ are significant with $p < 0.05$.}
    \label{tab:top_q_pairs}
\end{table}

\paragraph{Concerns about private influence track concerns about scale}
The strongest cross-section correlation is between \Q5-1 (there's too much focus on scale) and \Q1-1 (private firms have too much influence, $\rho_s = 0.43$).
Regarding scale, \Q5-1 was also moderately correlated with \Q6-7 (carbon footprint is a major concern, $\rho$ = 0.38).
These correlations suggest that NLP researchers
who see the influence of industry as problematic may hold this view in part because of concerns with the large-scale, compute-intensive research paradigm that is spearheaded largely by private firms.

\paragraph{Believing that LMs understand language is predictive of belief in AGI and the promise of scale}
Agreeing that text-only models can meaningfully ``understand'' language (\Q4-1) is predictive of several other views.
This who agree with \Q4-1 are more likely to believe that
scaling has moved us towards AGI (\Q3-2, $\rho_s = 0.42$) and can solve practically any NLP problem with existing techniques (\Q2-1, $\rho_s = 0.30$),
and less likely to believe that there is too much focus on scale (\Q5-1, $\rho_s = -0.35$), that linguistic structure is necessary to solve important NLP problems (\Q2-2, $\rho_s = -0.38$), or that we should do more to incorporate insights from domain sciences (\Q5-7, $\rho_s = -0.30$).
This suggests the existence of distinct `LM optimist' and `LM pessimist' positions, where people either believe that scaling up could solve most NLP problems and potentially lead to AGI,
or they think scaling is overprioritized, AGI is less likely, and we should be focusing more on seeking insights and methods from linguistics, cognitive science, or other domain sciences.
It is worth emphasizing, though, that none of our measured correlations here are extremely strong; people hold diverse sets of opinions and individuals cannot be cleanly split into these camps.

\subsection{Demographic Correlations}
\label{ssec:correlations-demographic}
\begin{figure}
    \centering
    \includegraphics[width=0.9\linewidth]{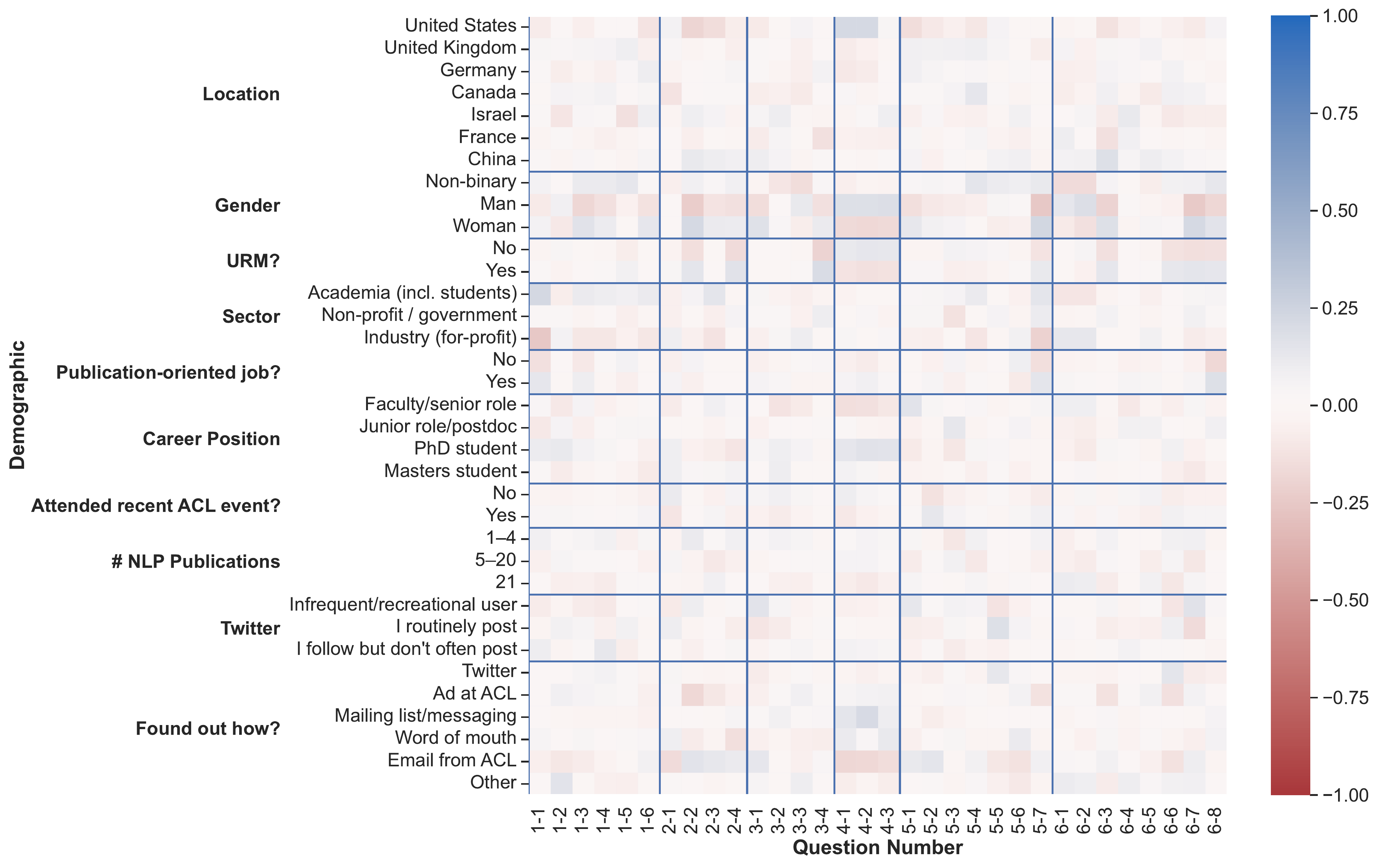}
    \caption{Spearman correlations between membership demographic groups and answers to agree/disagree questions. Lines separate survey sections and demographic variables. ``URM'' stands for under-represented minority. We only show demographic values with $>5$ respondents. Question numbers are given in \autoref{fig:big-table-o-questions}.}
    \label{fig:demog_to_q_spearman}
\end{figure}

Spearman correlations between demographic variables and agree/disagree questions are shown in \autoref{fig:demog_to_q_spearman}, with the top correlations by magnitude in \autoref{tab:top_demog_q}.
We rank agree/disagree answers as in \autoref{ssec:correlations-question}, and for
demographics, we treat each answer choice as a binary variable (1 if chosen by a respondent, 0 otherwise).
We exclude demographic values for which we have fewer than 5 responses.

\begin{table}[h]
    \centering
    \resizebox{\linewidth}{!}{
    \begin{tabular}{llr}
    \toprule
    Demographic & Question & $\rho_s$ \\
    \midrule
    Sector: Industry (for-profit) & \textbf{\Q1-1.} Private firms have too much influence. & \negate{-0.25} \\
    Gender: Woman & \textbf{\Q5-7.} We should incorporate more interdisciplinary insights. & \affirm{+0.24} \\
    Location: United States & \textbf{\Q4-2.} Multimodal models understand language. & \affirm{+0.22} \\
    Location: United States & \textbf{\Q4-1.} LMs understand language. & \affirm{+0.22} \\
    Sector: Academia (including students) & \textbf{\Q1-1.} Private firms have too much influence. & \affirm{+0.21} \\[0.4em]
    Gender: Man & \textbf{\Q6-2.} NLP's future net impact will be good. & \affirm{+0.21} \\
    Sector: Industry (for-profit) & \textbf{\Q5-7.} We should incorporate more interdisciplinary insights. & \negate{-0.21} \\
    Gender: Woman & \textbf{\Q6-7.} Carbon footprint is a major concern. & \affirm{+0.20} \\
    Gender: Woman & \textbf{\Q2-2.} Linguistic structure is necessary. & \affirm{+0.19} \\
    Under-represented Minority: Yes & \textbf{\Q3-4.} AI decisions could cause nuclear-level catastrophe. & \affirm{+0.19} \\
    \bottomrule
    \end{tabular}}
    \caption{Top 10 Spearman correlations ($\rho_s$) between membership in a demographic and answers to questions. Correlations $|\rho_s| > 0.11$ are statistically significant with $p < 0.05$.}
    \label{tab:top_demog_q}
\end{table}

Membership in demographic groups does not strongly correlate with answers to any questions in the survey.
The strongest correlation is $\rho_s = -0.25$~(\autoref{tab:top_demog_q}), smaller than the top 20 correlation coefficients between pairs of questions~(\autoref{tab:top_q_pairs}).
This suggests that there is a diversity of viewpoints in each demographic category, with more variation within demographics than between demographics.

Nonetheless, there were a few demographics that were particularly predictive of responses. For example, \textbf{Men and women answer many questions differently}, especially regarding ethics and belief in the value of interdisciplinary research.
In addition, \textbf{under-represented minorities are more likely to agree that AI could have catastrophic consequences this century}. Perhaps this reflects a worry that more powerful AI systems would have especially negative impacts on under-represented minorities, or that minorities would be negatively affected before society as a whole is significantly affected.

\subsection{Clustering}
\label{ssec:clustering}

To analyze the results beyond pairwise correlations, we identify clusters of opinions with principal component analysis (PCA).
To do this, we linearize the agree/disagree questions along $[-1, 1]$: $\{\Disagree \rightarrow -1, \WeaklyDisagree \rightarrow -0.5, \answerstyle{Other} \rightarrow 0, \WeaklyAgree \rightarrow 0.5, \Agree \rightarrow 1\}$,
where \answerstyle{Other} includes
\answerstyle{Question is ill-posed},
\answerstyle{Insufficiently informed on the issue},
and
\answerstyle{Prefer not to say}.
For demographics, we treat every answer choice as a 0/1 binary variable as in \autoref{ssec:correlations-demographic}.
This process gives us a total of 101 features for each respondent.

\begin{table}[t]
    \footnotesize
    \centering
    \begin{tabular}{lr|lr}
    \toprule
    \multicolumn{2}{c}{Scaling Maximalism (11.7\%)}  &
    \multicolumn{2}{c}{Concern about Fast Progress (5.5\%)} \\ %
    \midrule
    \textbf{\Q4-1.} LMs \affirm{do} understand language & +0.36 &
    \textbf{\Q3-1.} AGI \affirm{is} important & +0.48 \\ %
    
    \textbf{\Q4-2.} Multimodal models \affirm{do} understand & +0.29 &
    \textbf{\Q3-2.} We \affirm{are} stepping to AGI & +0.38 \\ %
    
    \textbf{\Q6-7.} Carbon \negate{isn't} a major concern & -0.29 &
    \textbf{\Q6-3.} Easy misuse \affirm{is} unethical   & +0.28 \\ %
    
    \textbf{\Q5-1.} There \negate{isn't} too much focus on scale & -0.28 &
    \textbf{\Q6-7.} Carbon \affirm{is} a major concern & +0.26 \\ %
    
    \textbf{\Q2-2.} Linguistic structure \negate{isn't} necessary & -0.26 &
    \textbf{\Q3-3.} Revolutionary change \affirm{is} plausible & +0.24 \\ %
    
    \midrule
    \multicolumn{2}{c}{Deep Learning Pessimism (5.2\%)} &
    \multicolumn{2}{c}{Jaded Empiricism (4.0\%)} \\ 
    \midrule
    \textbf{\Q5-5.} \affirm{Wrong} track (explainability) & +0.33 &
    \textbf{\Q6-6.} \negate{Not} unethical to predict psych. & -0.37 \\ 
    
    \textbf{\Q5-6.} \affirm{Wrong} track (interpretability) & +0.32 &
    \textbf{\Q1-5.} Most NLP \affirm{is} dubious science & +0.30 \\ 
    
    \textbf{\Q1-5.} Most NLP \affirm{is} dubious science & +0.28 &
    \textbf{\Q5-5.} \affirm{Wrong} track (explainability) & +0.27 \\
    
    \textbf{\Q3-4.} Catastrophic risk \affirm{is} plausible & +0.25 &
    \textbf{\Q5-6.} \affirm{Wrong} track (interpretability) & +0.24 \\
    
    \textbf{\Q6-2.} NLP \negate{isn't} good (future) & -0.23 & 
    \phantom{\textbf{\Q0-0.}}~\affirm{Has} 21+ Pubs. & +0.24 \\ 
    
    \bottomrule
    \end{tabular}
    \caption{The top four components from running PCA on survey responses and demographic data, with human-written cluster labels, percent variance explained in parentheses, and the questions/data with the highest magnitude associations per component. We \negate{negate} the statements with negative loadings so the statements for each component correspond to the set of beliefs that vary together.}
    \label{tab:pca_surveydemog}
\end{table}

We run PCA using \texttt{scikit-learn}
with 34 components, enough to explain 80\% of the variance in the data.
The variance in the data is fairly long-tailed,
with the first 8 components covering 41.1\% of the total variance and the remaining 26 explaining 38.9\% (with less than 3\% of variance explained by each component in the tail).
This indicates that perspectives among NLP researchers may be difficult to reduce to a small number of opposing camps (e.g., pro-scale and anti-scale) without missing a great deal of internal disagreement within those groups.

The top four principal components are shown in \autoref{tab:pca_surveydemog}.
The most prominent cluster of views in the data corresponds to the belief that we should (or shouldn't) be prioritizing large-scale modeling (``Scaling Maximalism''), explaining 11.7\% of the variance in the data, and aligning with the ``LM optimist'' perspective proposed in \autoref{ssec:correlations-question}.
Another prominent theme is concern with the pace of progress, characterized by a belief that we are making steps towards AGI and that it is an important concern for NLP researchers.
While other themes seem to appear in the components after that, no individual cluster of beliefs explains a large amount of the variance in the data, so these clusters are probably not very useful for directly reasoning about the beliefs of individuals, who are combinations of all principal components.

As found in \autoref{ssec:correlations-demographic}, demographic features are not as explanatory of the data as the agree/disagree questions are.
Accordingly, they are not among the questions most strongly associated with the top clusters.
The exception is the fourth component (``Jaded Empiricism''), for which one of the strongest associations is having a large number of publications, though this component only explains 4.0\% of the total variance.

\section{Related Work}

This work is directly inspired by the 2009 PhilPapers Surveys~\citep{bourget2014what}, a survey and metasurvey of professional philosophers designed to discover and compare both their philosophical beliefs and their sociological beliefs about the philosophy community.
More generally, introspective surveys to get a sense of the demographics and opinions of a field are common in many disciplines, such as
economics \citep[][\textit{inter alia}]{fuchs1998economists,illge2009matter,frey2010economics} and
physics \citep{sivasundaram2016surveying}.
We design our survey to fill a similar role for NLP,
with a special focus on controversies where the field's perception of its views may not match reality, due for example to the influence of social media or a small number of dominant voices.

The NLP Community Metasurvey overlaps slightly with other surveys of the NLP community, which are typically organized by ACL committees and focused on specific, timely issues related to the organization policy or logistics.
Issues covered by such surveys include
preprint posting practices~\citep{survey_preprints},
experiences with rolling review infrastructure~\citep{survey_reviewing},
ethical stances~\citep{fort2016yes,survey_ethics},
attitudes towards energy use and environmental impacts~\citep{survey_efficiency},
and language diversity in NLP research~\citep{survey_language_diversity}.
Instead of focusing on a specific issue, our aim with the NLP Community Survey is to provide a broad sense of the distribution of views (and meta-views) held by NLP researchers. To do this, we cover a range of topics with existing debate in the literature (see the subsections of \autoref{sec:results} for references).
This is related to the the NLP Scholar project, which analyzes broad demographic \citep{mohammad2020gender,mohammad2019nlpscholar}, publication \citep{mohammad2020data}, and citation \citep{mohammad2020citations} trends in NLP and their interplay.

\section{Discussion}
\label{sec:conclusion}

With the NLP Community Metasurvey, we have made concrete numbers of many contentious issues in NLP: the necessity of expert-designed inductive bias, the importance of AGI, whether language models understand language, and more.
Perhaps more interestingly, we have also made concrete numbers of the community's \textit{impressions} of these controversies, in some cases confirming what we already believe and in others producing surprises.
For example, the idea that mere scale will solve most of NLP is much less controversial (and much less believed) than it is thought to be, and NLP researchers unexpectedly agree that we should do more to incorporate insights and methods from domain sciences, and that we should prioritize problem formulation and task design.
Interestingly, very few of the issues we ask about (only the necessity of linguistic structure and expert-designed inductive bias, \Q2-2 and \Q2-3) are noticeably more controversial than respondents expected them to be.
This could be due to biases from the amplification of controversy (e.g., in social media), or it could just reflect mundane biases in respondent predictions, e.g., regression towards the middle of the range ($\sim$50\%) under uncertainty.

There are other biases to keep in mind when interpreting our results: people in the United States are overrepresented in our respondent population relative to ACL members as a whole, and senior researchers and academics are probably overrepresented as well~(\autoref{app:demographics}),
not to mention unmeasured population biases based in the personal networks of the authors.
Many of the questions have multiple possible interpretations, many rely on vague terms like ``plausible'' or ``major concern,'' and some rely on comparisons to reference points such as the Industrial Revolution (\Q3-3) or ``specific, non-trivial results from...~linguistics or cognitive science'' (\Q2-4) which may carry different implications for different readers.
Given these issues, it is probably reasonable to view the answers to the questions on this survey as reflecting \textit{something between objective beliefs and signaling behavior}.
Agreement or disagreement with a particular statement may indicate where a respondent believes they stand relative to ``received wisdom,'' which would determine what statements would be worth asserting in the context of the status quo;
or, a response could be driven by identification with (or rejection of) an already-known ideological camp that the statement is taken to refer to.
While these issues affect the way we should interpret the absolute numbers in our results,
they should apply equally to the meta-questions, 
so we believe it is meaningful to compare the survey's actual and predicted results as a way of discovering 
false sociological beliefs.

We hope the results of the NLP Community Metasurvey can help us update our sociological beliefs to closer match reality, creating common ground for fruitful and productive discourse among NLP researchers as we confront these issues in the course of our work.

\subsubsection*{Acknowledgments}
We thank the 480 respondents who completed the survey, as well as (especially) our 26 pilot testers and colleagues who provided feedback on the survey.

This project has benefited from financial support to SB by Eric and Wendy Schmidt (made by recommendation of the Schmidt Futures program) and Apple. This material is based upon work supported by the National Science Foundation under Grant Nos.\ 1746891, 1922658 and 2046556. Any opinions, findings, and conclusions or recommendations expressed in this material are those of the author(s) and do not necessarily reflect the views of the National Science Foundation. 

\bibliographystyle{iclr/iclr2021_conference}
\bibliography{references}

\appendix

\section{Overview of Responses}
\label{app:responses-overview}
\begin{figure}[htbp]
\begin{center}
\includegraphics[width=\textwidth]{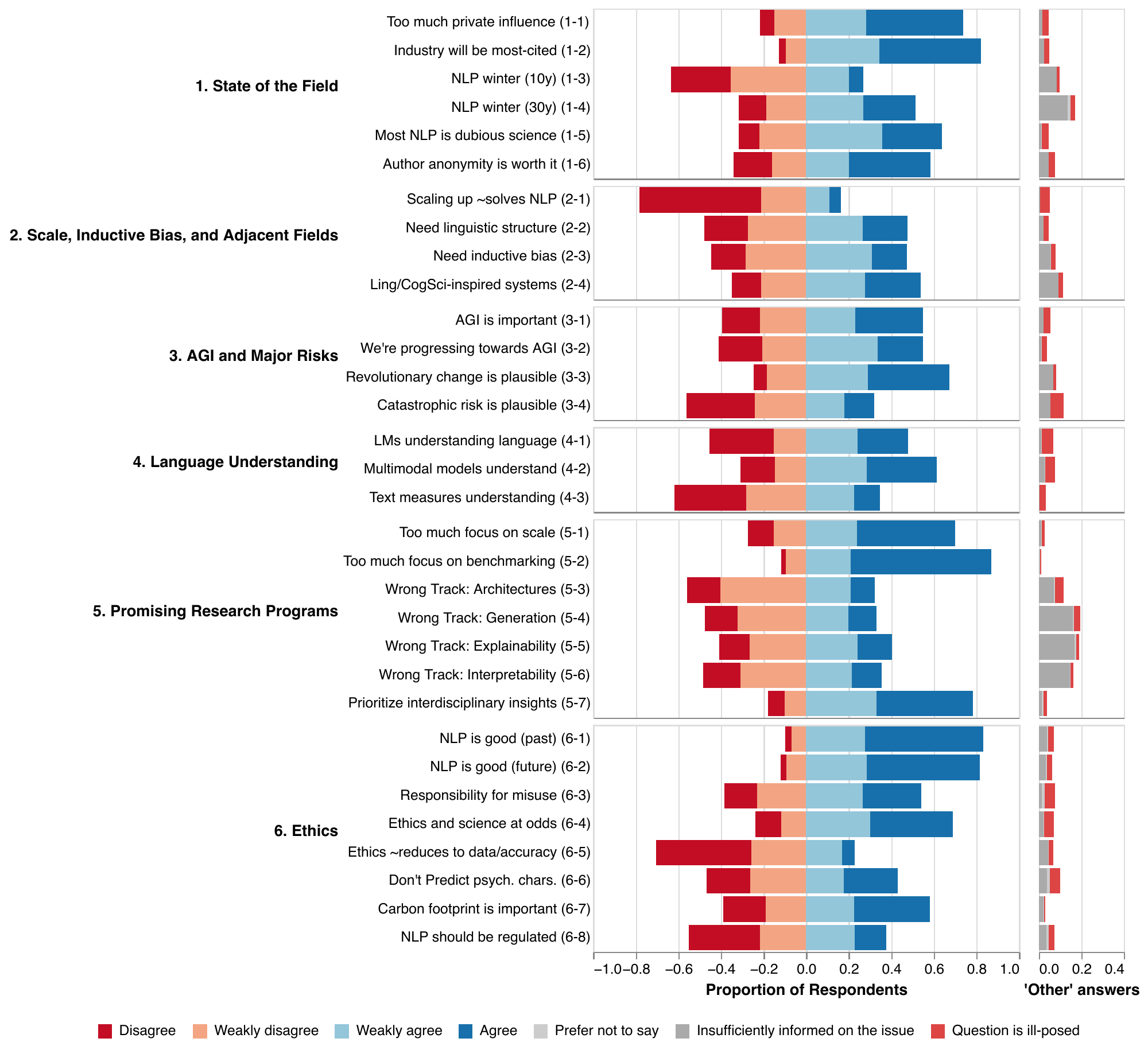}
\end{center}
\caption{Full overview of responses. \label{fig:results-overview}}
\end{figure}

A full overview of the responses to agree/disagree questions is shown in \autoref{fig:results-overview}.
The \answerstyle{Other} answers were fairly uncommon---never above 20\% of all answers---and the most frequent one was \answerstyle{Insufficiently informed on the issue}, which many respondents gave for the questions about an NLP winter (\Q1-3, \Q1-4) and the ``wrong track'' questions (\Q5-\{3--6\}).

\begin{figure}[htbp]
\begin{center}
\includegraphics[width=\textwidth]{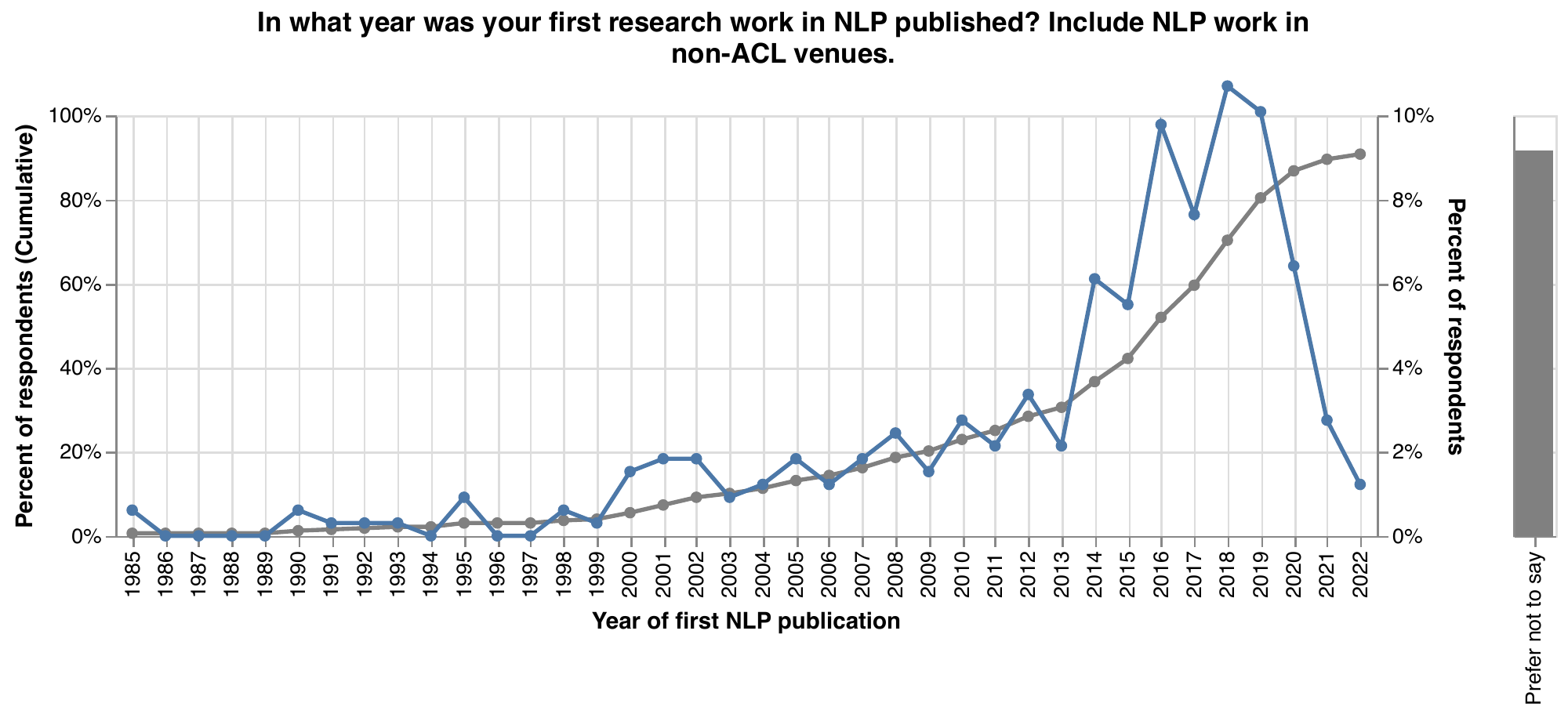}
\end{center}
\caption{Respondents' year of first NLP publication. \label{fig:demog-firstpub}}
\end{figure}

\section{Detailed Demographics}
\label{app:demographics}
\begin{figure}[htbp]
\begin{center}
\includegraphics[width=\textwidth]{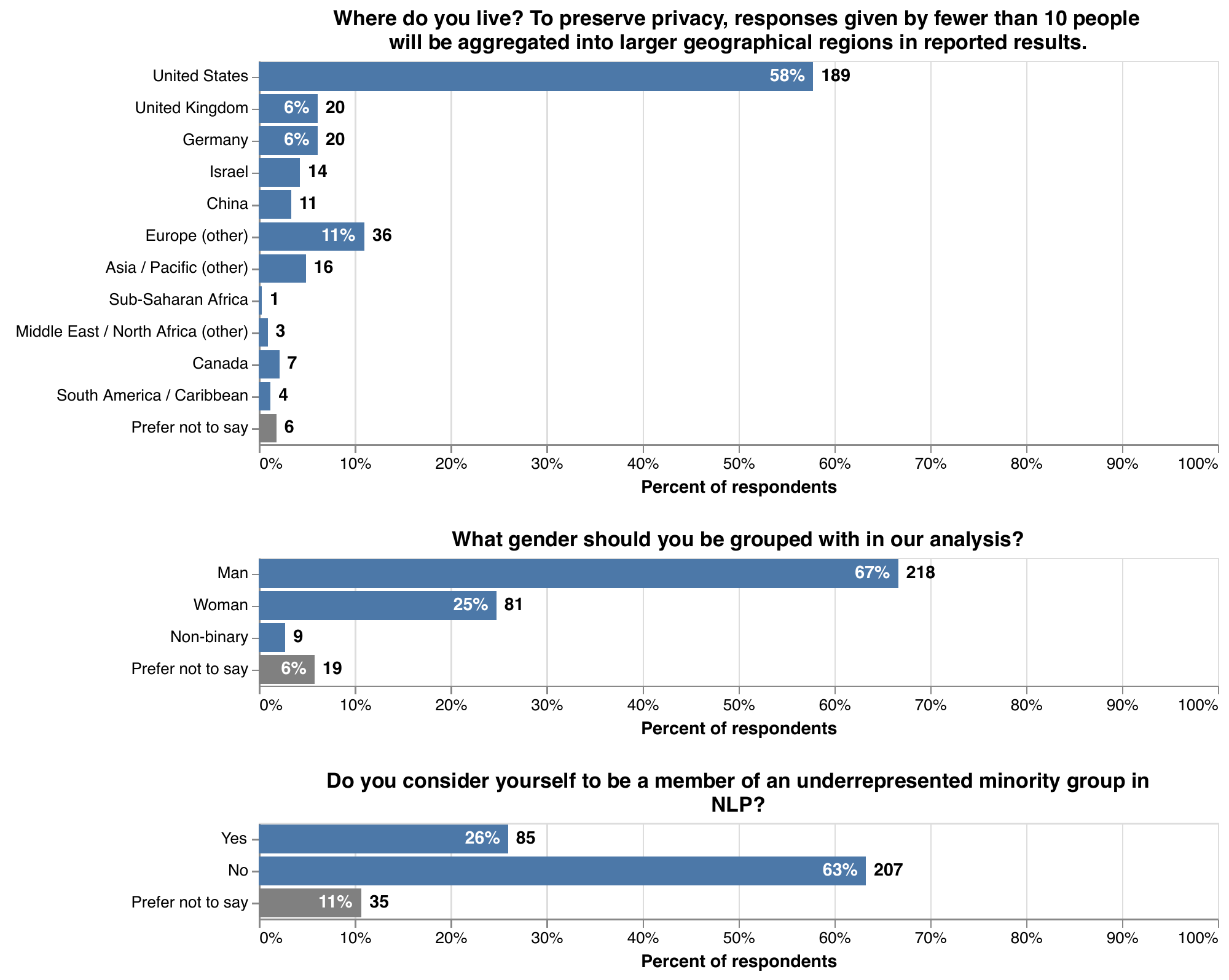}
\end{center}
\caption{Basic demographics. \label{fig:demog-basic}}
\end{figure}

\begin{figure}[htbp]
\begin{center}
\includegraphics[width=\textwidth]{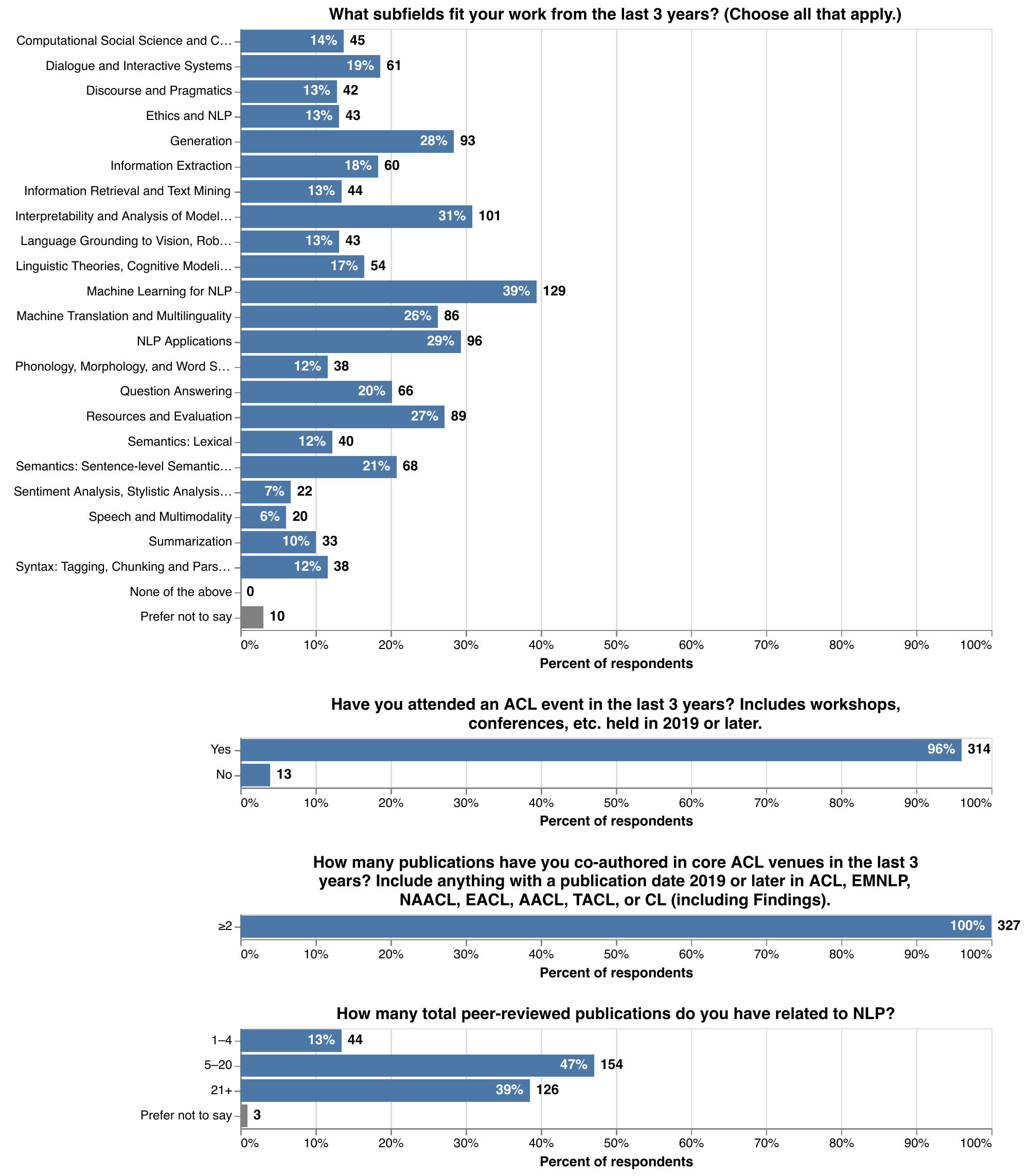}
\end{center}
\caption{Respondents' research activities. \label{fig:demog-research}}
\end{figure}

\begin{figure}[htbp]
\begin{center}
\includegraphics[width=\textwidth]{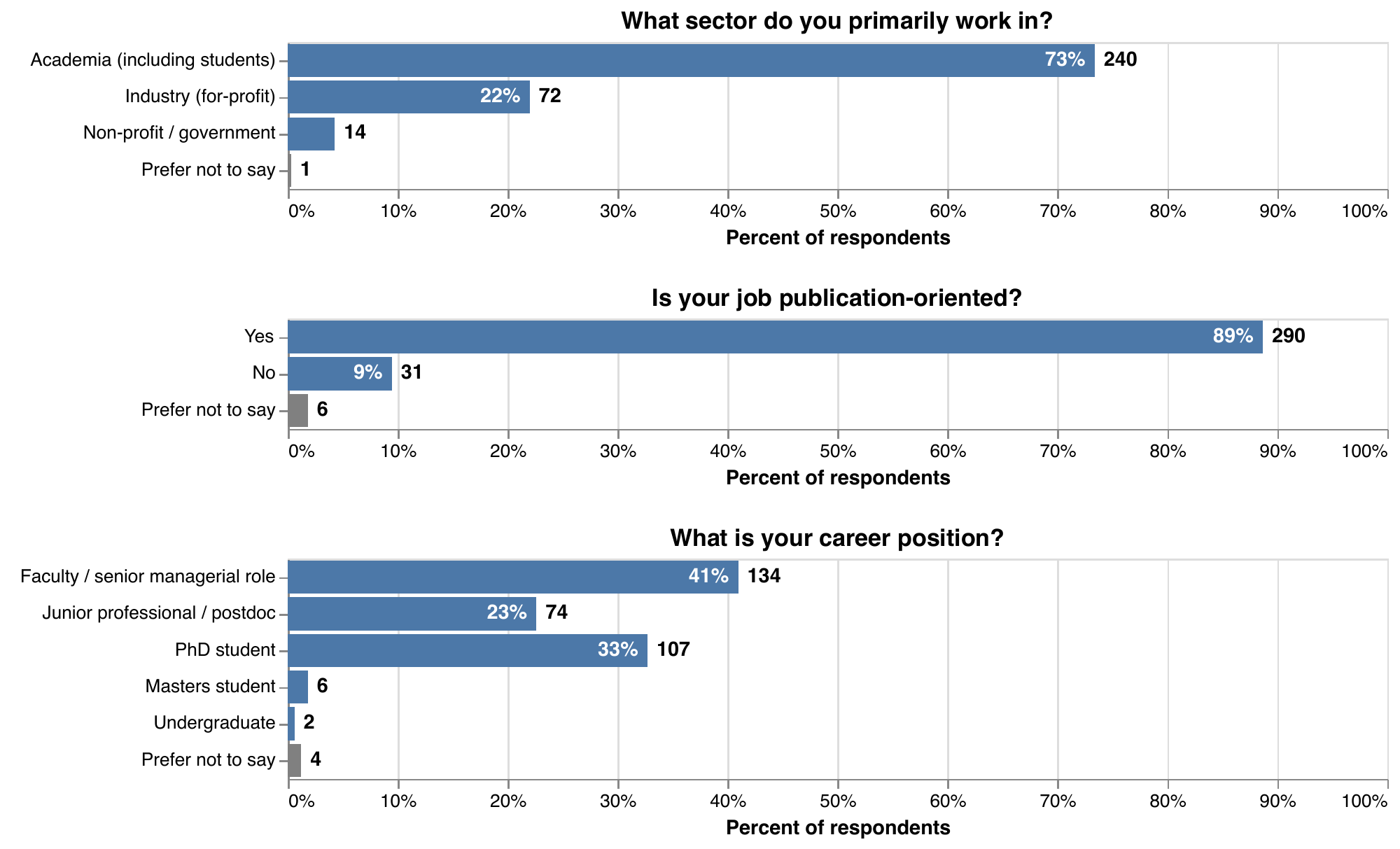}
\end{center}
\caption{Career demographics. \label{fig:demog-career}}
\end{figure}

\begin{figure}[htbp]
\begin{center}
\includegraphics[width=\textwidth]{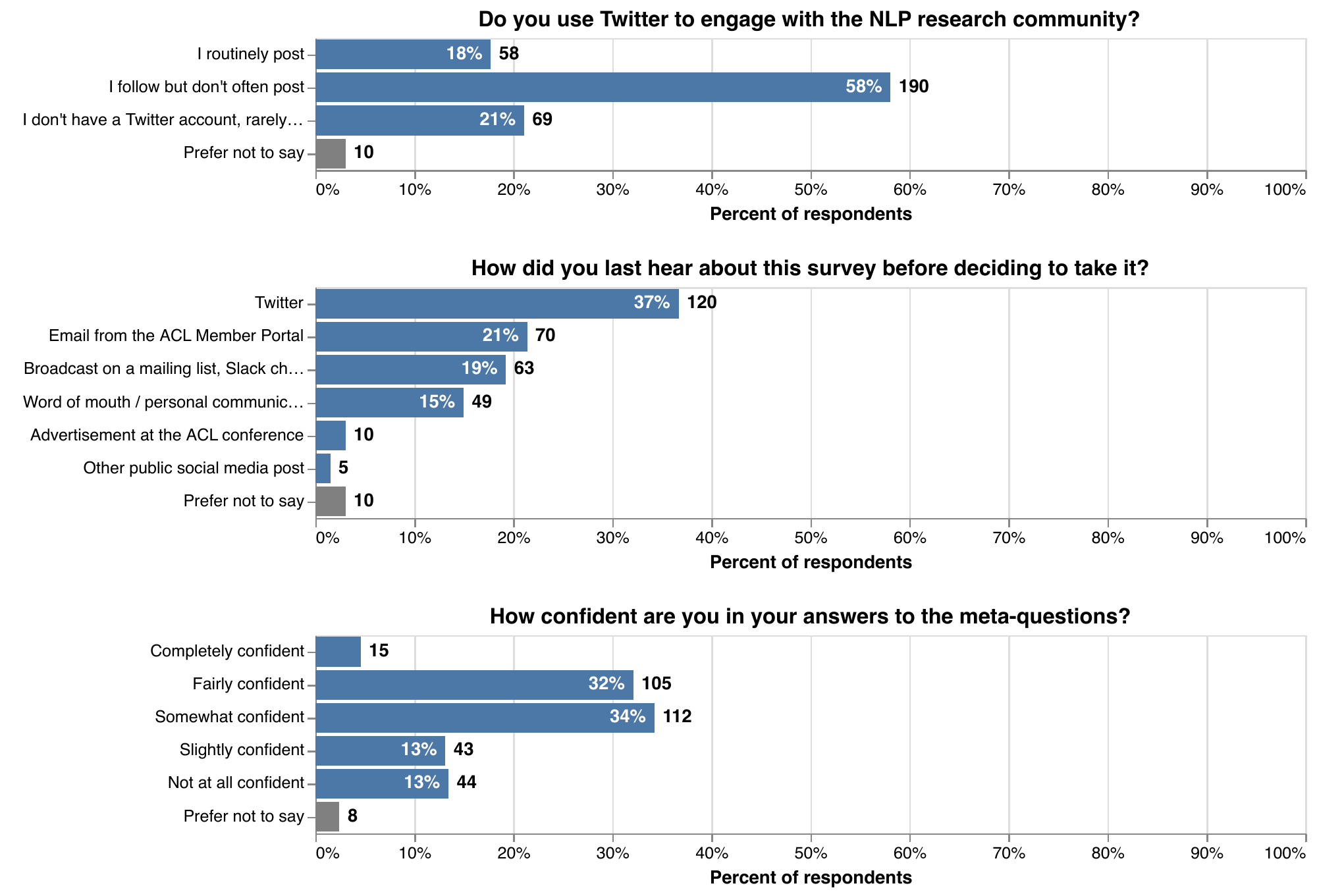}
\end{center}
\caption{Other information provided by respondents. \label{fig:demog-other}}
\end{figure}

Figures \ref{fig:demog-basic}--\ref{fig:demog-other} show full results for the demographics questions. Results are restricted to those in the target demographic, i.e., with at least 2 *CL publications in the last three years. Numeric labels for percentages below 5\% are omitted from the charts for space.

\section{Pilot Testing}
\label{app:pilots}
The first author conducted 6 pilot studies with about 26 different participants from Computer Science and Linguistics departments, mostly based in the United States, during the months of February and March of 2022. After pilot participants took the survey, they were asked for feedback in a group Zoom call. Participants were asked about any questions they perceived as leading, reasons they might have refused to answer questions, reasons they might have wanted to stop taking the survey in the middle, and whether the purpose of the survey was clear, etc.
The survey instructions and question wording were updated in accordance with their feedback.

\section{Data Cleaning and Postprocessing}
\label{app:cleaning}

\begin{figure}[htbp]
\begin{center}
\includegraphics[width=\textwidth]{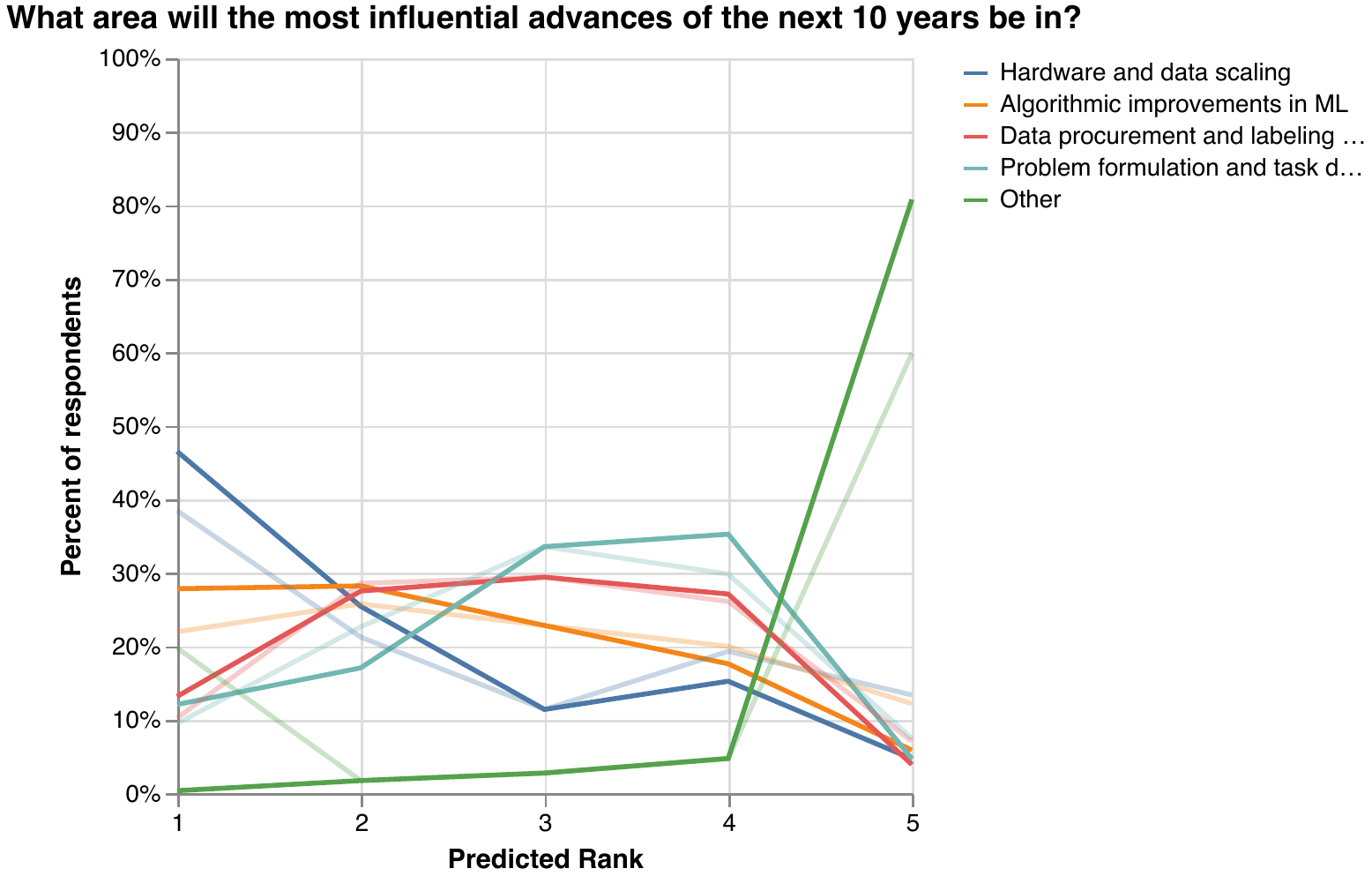}
\end{center}
\caption{We postprocessed the predicted rankings for likely sources of future advances by reversing the rankings given by all respondents who placed ``Other'' first (20\% of respondents). The rankings for the unadjusted data are shown in the faded lines; besides the change to the ``Other'' line, overall trends are the same. \label{fig:advances-adjustment}}
\end{figure}

\paragraph{Likely Sources of Future Advances}
As mentioned in \autoref{ssec:future}, 20\% of respondents who answered the meta-question on likely sources of future advances ranked ``Other'' as the most common answer to the question. We assume these were mistakes, since it is unlikely that people would think a plurality of respondents would reject all of the (fairly broad) provided options.
We take this, then, to mean the rankings provided by these respondents were probably reversed, with 5 being the most common and 1 the least common.
So we reverse the rankings provided by these respondents for the purposes of analysis.
Besides changing the ``Other'' statistics, this does not seem to have a noticeable effect on overall trends (\autoref{fig:advances-adjustment}).

\clearpage
\subsection{Survey Instructions}
\label{app:instructions}

\begin{figure}[htbp]
\begin{center}
\includegraphics[width=0.48\textwidth]{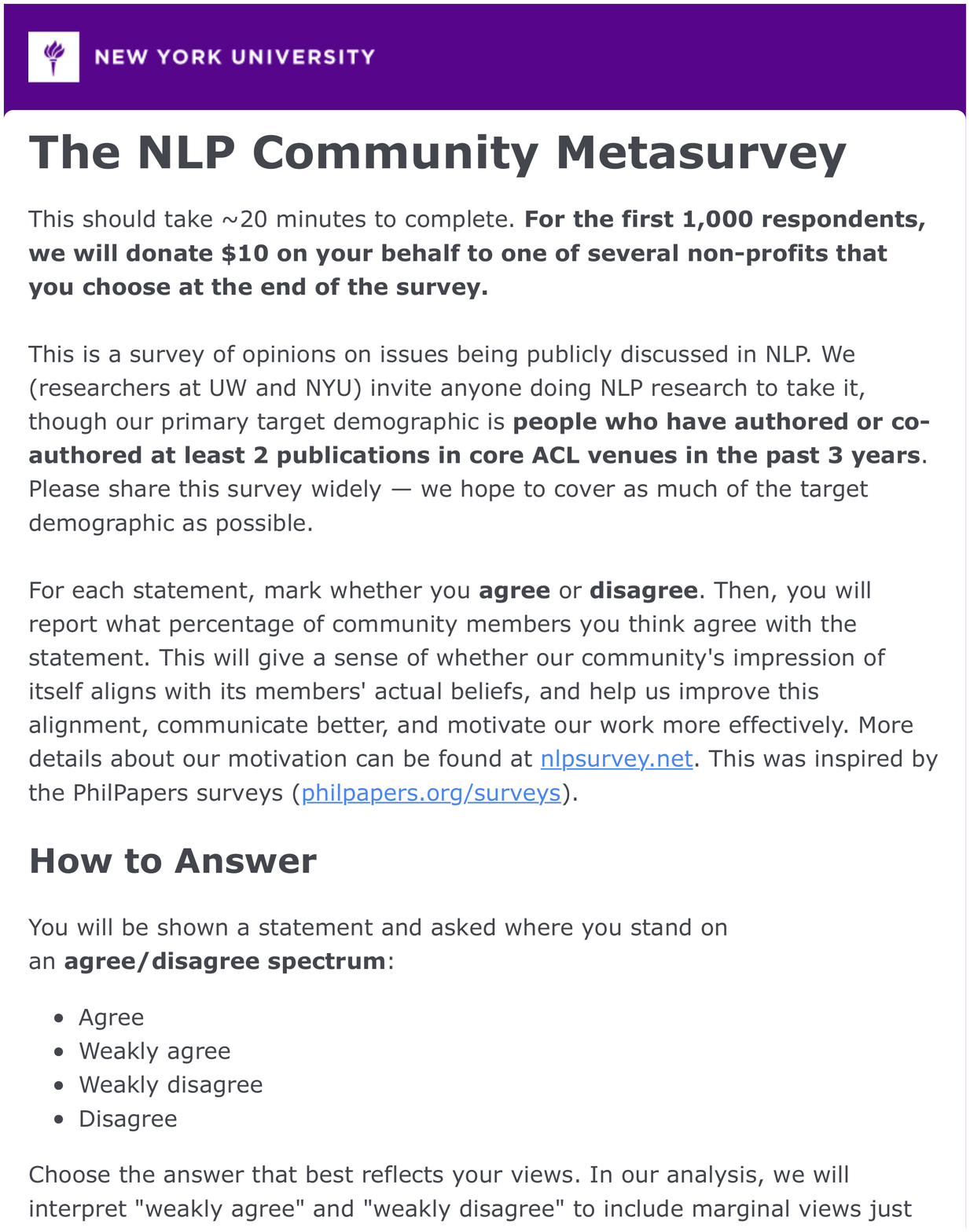}
\includegraphics[width=0.48\textwidth]{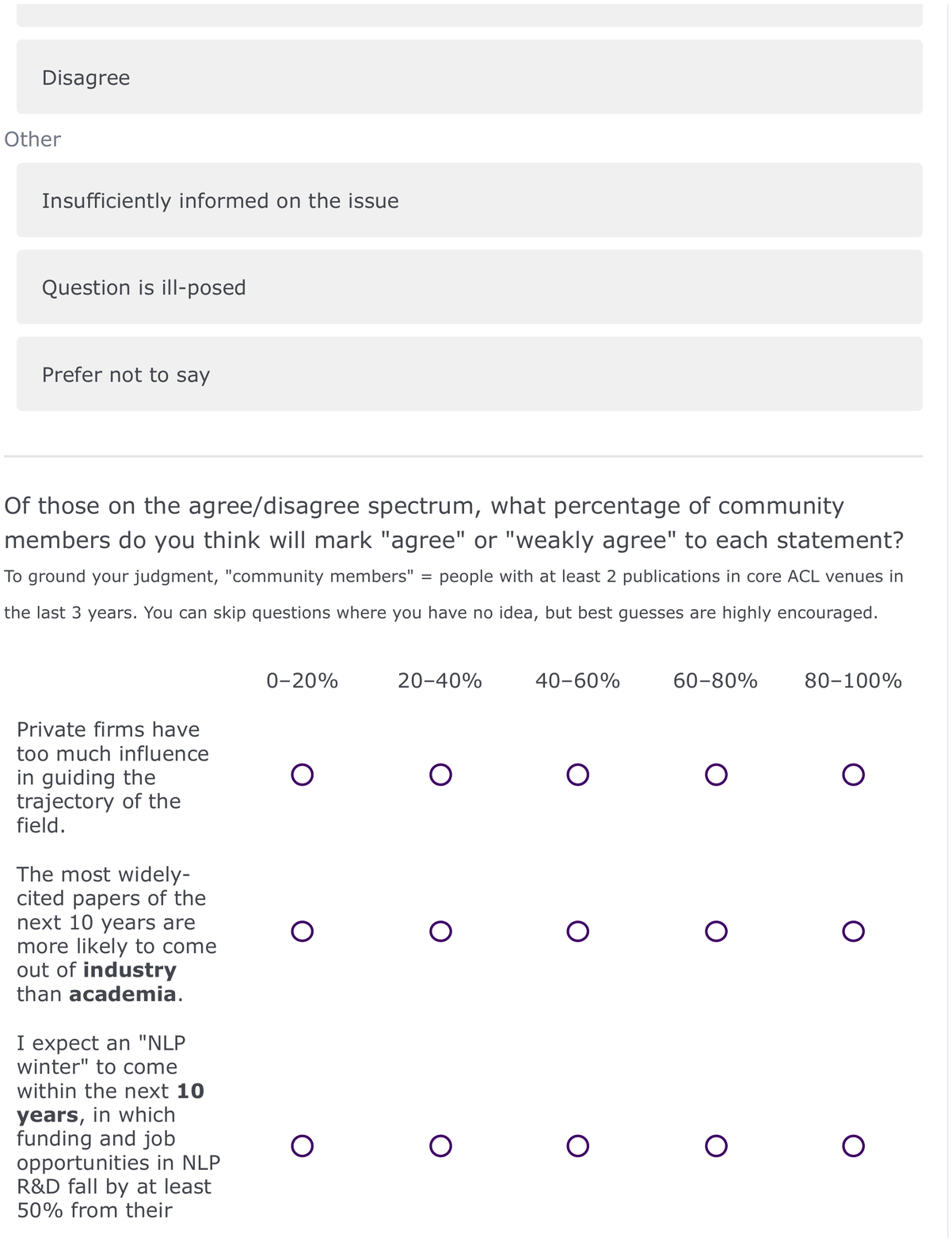}
\includegraphics[width=0.48\textwidth]{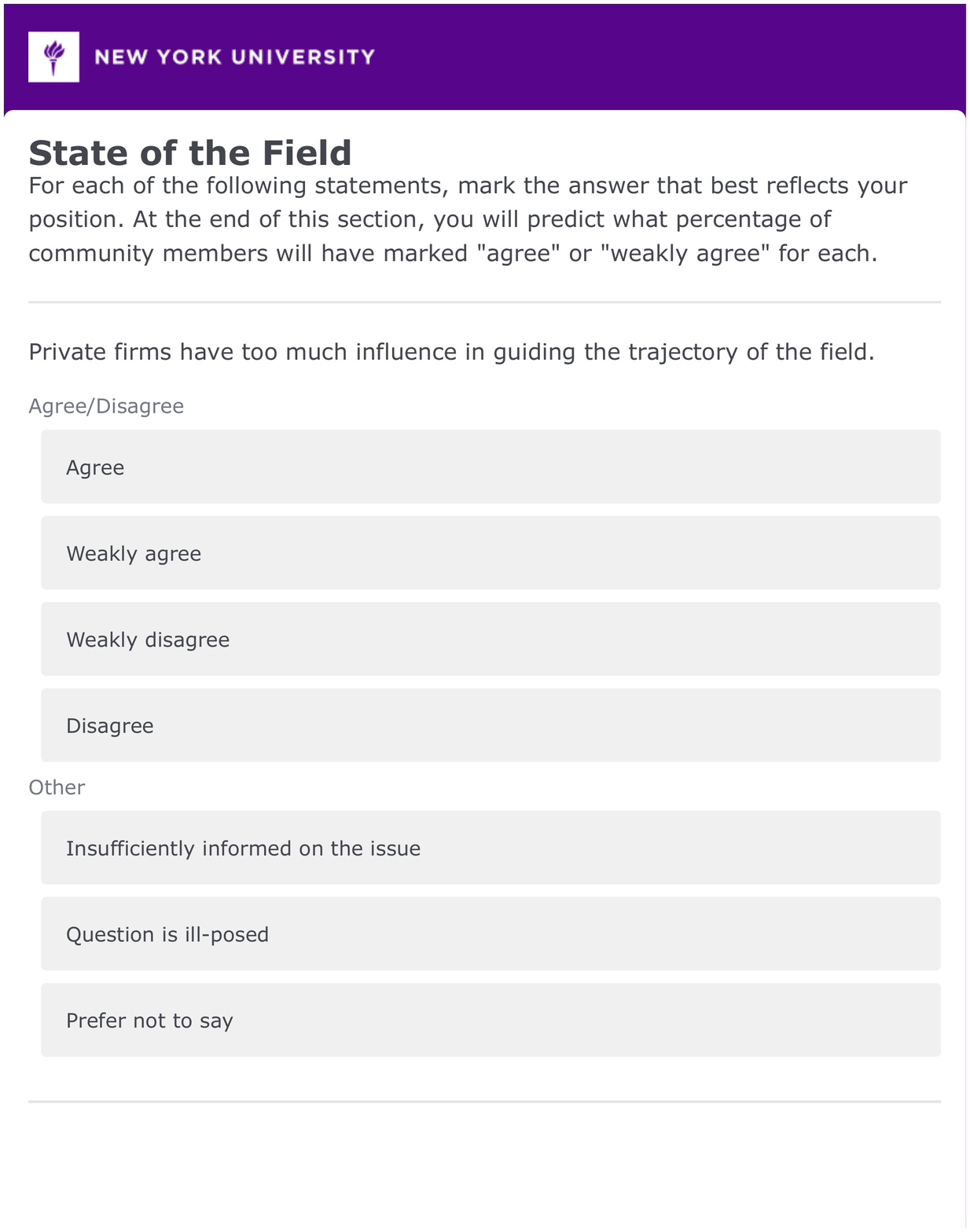}
\includegraphics[width=0.48\textwidth]{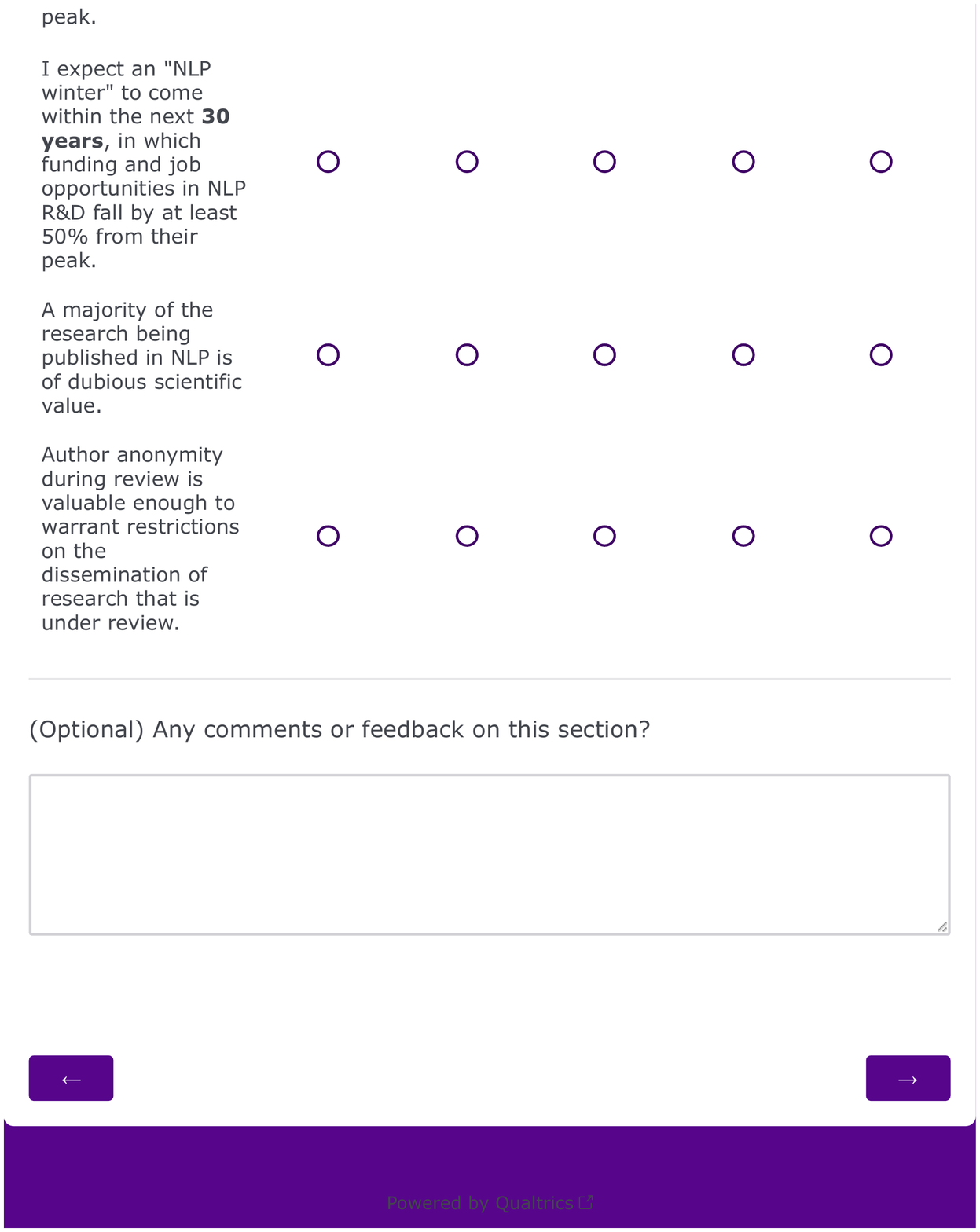}
\end{center}
\caption{How the survey looks to respondents in a web browser. \label{fig:survey-screenshots}}
\end{figure}

The survey instructions are reproduced in full below, and examples of how it looks in the browser are given in \autoref{fig:survey-screenshots}.

\noindent\fbox{%
    \parbox{\textwidth}{%
    
    \small

\subsubsection*{The NLP Community Metasurvey}
 
\vspace{0.5em}\noindent
This should take $\sim$20 minutes to complete. \textbf{For the first 1,000 respondents, we will donate \$10 on your behalf to one of several non-profits that you choose at the end of the survey.}
 
\vspace{0.5em}\noindent
This is a survey of opinions on issues being publicly discussed in NLP. We (researchers at UW and NYU) invite anyone doing NLP research to take it, though our primary target demographic is \textbf{people who have authored or co-authored at least 2 publications in core ACL venues in the past 3 years.} Please share this survey widely --- we hope to cover as much of the target demographic as possible.

\vspace{0.5em}\noindent
For each statement, mark whether you \textbf{agree} or \textbf{disagree}. Then, you will report what percentage of community members you think agree with the statement. This will give a sense of whether our community's impression of itself aligns with its members' actual beliefs, and help us improve this alignment, communicate better, and motivate our work more effectively. More details about our motivation can be found at \url{nlpsurvey.net}. This was inspired by the PhilPapers surveys (\url{philpapers.org/surveys}).
 
\vspace{0.5em}
\paragraph{How to Answer}
You will be shown a statement and asked where you stand on an \textbf{agree/disagree spectrum}:
\begin{itemize}
\item Agree
\item Weakly agree
\item Weakly disagree
\item Disagree
\end{itemize}
Choose the answer that best reflects your views. In our analysis, we will interpret ``weakly agree'' and ``weakly disagree'' to include marginal views just barely agreeing or disagreeing (e.g., ``depends, leaning positive/negative''). However, in case you cannot place yourself on either side of an issue, there will also be three non-answer options:
\begin{itemize}
\item \textbf{Insufficiently informed on the issue}: You don't understand the statement or its subject matter well enough to form an opinion.
\item \textbf{Question is ill-posed}: You reject the distinction between agreeing and disagreeing, or don't think the statement admits any coherent interpretation.
\item \textbf{Prefer not to say}: You don't feel comfortable providing any of the other answer choices.
\end{itemize}
If you pick ``question is ill-posed'' or ``prefer not to say,'' we would appreciate (optional) feedback at the end of the respective section explaining your reasons so we can better interpret the results.
 
\vspace{0.5em}
\paragraph{Meta-Questions}

At the end of each section, you'll be asked to predict what proportion of people on the agree/disagree spectrum will answer \textbf{either ``agree'' or ``weakly agree''} to each question.

For the purpose of these questions, please predict relative to the target demographic: people with at least 2 publications in core ACL venues in the last 3 years. For our purposes, core venues are ACL, EMNLP, NAACL, EACL, AACL, TACL, and CL (including Findings). By our count from the ACL Anthology, this includes approximately 5,650 people.

Even if you don't have a strong sense of the community's stance, give your best guess (unless you really feel like you have no priors, in which case you can skip these questions). At the end of the survey, you will rate your overall confidence in the meta-survey questions so we can account for it in our analysis.
 
\vspace{0.5em}
\paragraph{Privacy}

Your responses are anonymous and individual responses will not be released publicly. You will have the option to de-anonymize yourself at the end; this will help us audit the results and follow up with you after the survey, but will not be released publicly or shared with anyone without your permission. In accordance with the General Data Protection Regulation (GDPR), all survey respondents have rights over their personally-identifiable information. This form (\url{nlpsurvey.net/gdpr.pdf}) outlines those rights and how to exercise them. A list of the people who will have access to the non-anonymized data is available at \url{nlpsurvey.net/about}.

\vspace{0.5em}\noindent
You can reach us with questions and concerns at \texttt{nlp-metasurvey-admin@nyu.edu}.
    }%
}

\end{document}